\documentclass{article}

\usepackage{PRIMEarxiv}

\usepackage[utf8]{inputenc} 
\usepackage[T1]{fontenc}    
\usepackage{hyperref}       
\usepackage{url}            
\usepackage{booktabs}       
\usepackage{amsfonts}       
\usepackage{nicefrac}       
\usepackage{microtype}      
\usepackage{lipsum}
\usepackage{fancyhdr}       
\usepackage{tabularx}
\usepackage{graphicx}
\usepackage{xcolor}
\usepackage{lscape}
\usepackage{multirow}
\usepackage{adjustbox}
\usepackage{rotating}
\usepackage{longtable}
\usepackage{amsmath}
\usepackage{float}
\usepackage{changepage}
\graphicspath{{media/}}     

\title{A Comparative Study of Real-Time Implementable Cooperative Aerial Manipulation Systems
}

\author{
  Stamatina C. Barakou\\
  School of Electrical and Computer Engineering \\
  National Technical University of Athens \\
  Athens, Greece \\
  \texttt{matbarakou@gmail.com} \\
   \And
  Costas S. Tzafestas \\
  School of Electrical and Computer Engineering \\
  National Technical University of Athens \\
  Athens, Greece \\
  \texttt{ktzaf@cs.ntua.gr} \\
   \And
  Kimon P. Valavanis \\
  Department of Electrical and Computer Engineering \\
  University of Denver, Denver, CO 80210, USA \\
  \texttt{kimon.valavanis@du.edu} \\
}

\begin{document}
\maketitle

\begin{abstract}
This survey paper focuses on quadrotor- and multirotor- based cooperative aerial manipulation. Emphasis is first given on comparing and evaluating prototype systems that have been implemented and tested in real-time in diverse application environments. Underlying modeling and control approaches are also discussed and compared. The outcome of the survey allows for understanding the motivation and rationale to develop such systems, their applicability and implementability in diverse applications and also challenges that need to be addressed and overcome. Moreover, the survey provides a guide to develop the next generation of prototype systems based on preferred characteristics, functionality, operability and application domain.
\end{abstract}


\section{Introduction}

Research and development in Unmanned Aerial Vehicles (UAVs) or Unmanned Aircraft Systems (UAS) has witnessed unprecedented scientific and commercial interest and growth, particularly during the last two decades. Although military applications dominated the global market for years, interest in using UAVs in civil and public domains increases exponentially, worldwide, albeit challenges related to integrating unmanned aviation into the national airspace. Sample applications include, but are not limited to, surveillance \cite{9}, search and rescue \cite{8}, aerial photography \cite{3}, fire monitoring \cite{7}, agriculture \cite{6}, and aerial delivery \cite{25}. The listed applications refer to solely passive tasks, that is, tasks in which no UAV interaction with the environment is needed. However, contact with the environment is required in  industrial and maintenance applications like bridge inspection, water damn inspection, high-voltage transmission line inspection \cite{1}, assembly tasks \cite{EU} or construction \cite{4}. This requirement is in addition to, obviously, navigation and control, stability consideration and accurate manipulation, to say the least.   

When focusing on aerial manipulation \cite{14}, based on requirement analysis, specifications and applications, different multirotor designs and configurations with attached robotic arms offer alternative solutions to perform tasks under full teleoperation, autonomously or semi-autonomously; these tasks are considered to be dangerous for human operators and also costly \cite{1}. The multirotor configuration and its structure allows for hovering and for performing complex maneuvers while reaching high altitudes fast enough, thus, facilitating completion of aerial manipulation tasks in diverse environments. 

Aerial manipulation has challenges that need to be addressed and overcome, like stability in the presence of forces/torques from the attached manipulator or payload, and because of complicated dynamics due to under-actuation and payload weight constraints. Many challenges have already been considered in the literature \cite{5}. But, there still exist challenges related to cooperative aerial manipulation, where two or more aerial platforms (multirotors) manipulate or transport a payload that is too heavy or too big for a single multirotor to carry \cite{14}. 

Cooperative aerial manipulation may impact human workforce in highly repetitive and heavy work, or in remote and dangerous areas, which are inaccessible to humans. Mission complexity, in addition to the overall multirotor system complexity, is higher due to the required task(s) distribution, planning, coordination and cooperation of each arm \cite{2}. 

When considering hard real-time (or almost hard real-time) implementable cooperative aerial manipulation multirotor systems, specific  challenges that need to be addressed include, among others:
\begin{itemize}
    \item Ability to navigate in uncertain environments (i.e., poor GPS or GPS denied areas). Real-time SLAM, adaptation to unforeseen events, dynamic collision avoidance, planning and re-planning and safe flight based on onboard sensor suites are minimum requirements to safe navigation.         
    \item Power and energy requirements (battery consumption), algorithm computational complexity, execution time, effective payload, maximum takeoff weight (MTOW) must be considered as they affect flight-time, range and endurance. 
    \item Robust communication among platforms is essential, with minimum or no delays (due to latency, transmission loss, etc.), as well as minimum or no down-time. 
    \item System stability must be tackled since, as stated in \cite{101}, wind gusts, random wind profiles, aerodynamic perturbations, induced, parasitic, and other types of drag, may affect accurate manipulation and navigation.
    \item On the regulation front, regulations dictate that a certified pilot must operate a UAV - this refers to teleoperation as an alternative to semi-autonomous or autonomous functionality (for safety reasons). Obviously, autonomous flight may be needed in certain applications. 
\end{itemize}

This survey centers around reviewing real-time implementable cooperative aerial manipulation systems. It presents modeling and control approaches and it summarizes their advantages and limitations. Throughout the survey, the term 'system' refers to, and includes, the multirotor platform(s), attached robotic manipulator/arm, navigation and control approaches, and grasping capabilities.

The rest of the paper is organized as follows. Section 2 summarizes the Search Method that has been followed throughout the survey. Section 3 provides a concise summary of aerial manipulation. Section 4 classifies different approaches to cooperative aerial manipulation including: cooperative cable-suspended manipulation; multiple multirotors transporting an object tethered with cables; aerial manipulation with multi-DOF arms; multirotors with robotic arms; ground-air manipulation; aerial robots cooperating with ground vehicles; manipulation of flexible objects; multirotors carrying flexible objects instead of rigid ones and rigidly object attached multirotors. Section 5 discusses modelling approaches of cooperative aerial manipulation, that is, Newton-Euler, Euler-Lagrange and also combined methods. Section 6 presents and classifies the adopted, derived and used control techniques (controller design techniques) in four categories; control of plain manipulation/transportation; control in the presence of wind; vision-based and teleoperation. Section 7 offers discussion, it summarizes results and concludes the survey. It is stated that Sections 3 and 4 are enhanced versions of the authors' previous work \cite{125} that has been published in 2023 International Conference on Unmanned Aircraft Systems (ICUAS), while Sections 5 and 6 are completely new.

\section{Search Method}

The search method that has been followed to identify related articles and previous reviews on the subject of "cooperative aerial manipulation" is shown in Table \ref{tab:my-table}. Published research papers with experimental demonstrations, as well as experiments with simulations and numerical results, are also included and classified. Although the complexity of such systems and, mostly, any experimental validation, are both crucial as they provide proof-of-concept demonstration in terms of applicability and implementability, it is also important for the research community to be aware of all developments in cooperative aerial manipulation, of the research maturity in the field, and of all developments from the early conceptual design stage to actual real-time implementation and testing. The included papers have been reviewed for originality and technical quality, while considering their relevance with the previously mentioned challenges. Note that thesis and unpublished research papers (that have been found through internet search) are not included in the survey.

\begin{table}[h]
\centering
\footnotesize
\caption{Search criteria}
\label{tab:my-table}
\resizebox{!}{!}{%
\begin{tabular}{ll}
\toprule
\textbf{Criteria} & \textbf{Data} \\ \midrule
\textit{Scientific Database} & \begin{tabular}[c]{@{}l@{}}IEEE   Xplore, Google Scholar, \\ Science Direct, Engineering Village, \\ arXiv, manual search\end{tabular} \\ \midrule
\textit{Publication Period} & From 2010 to October 2023 \\ \midrule
\textit{Keywords} & \begin{tabular}[c]{@{}l@{}}(“aerial” OR “cooperative aerial” OR “survey” OR “review”) \\ AND (“manipulation” OR “transportation” OR “load transportation”)\end{tabular} \\\bottomrule 
\end{tabular}%
}
\end{table}

\section{Literature Review on Aerial Manipulation}

The survey on aerial manipulation focuses mostly on published research during the last decade. As mentioned in \cite{14} and \cite{101}, aerial manipulation has mostly been demonstrated in indoors environments. Representative research that includes testing in outdoors settings and in real world scenarios is found in \cite{88}. In \cite{14}, authors state that aerial manipulation may be divided in three categories according to the specific platform, manipulation arms, navigation, perception and planning. 

\begin{figure}[thpb]
    \includegraphics[width=0.33\linewidth,height=30mm]{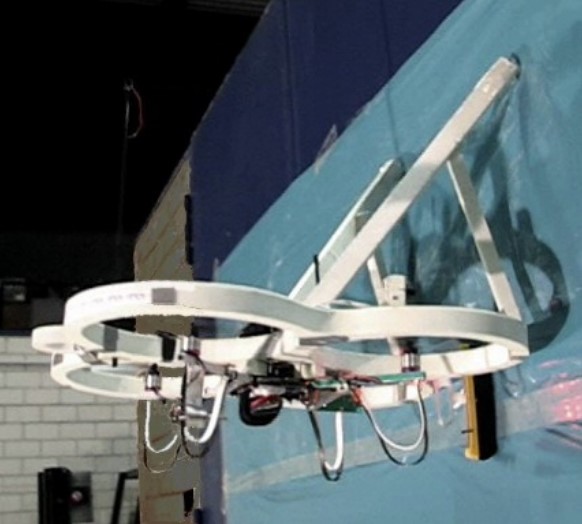}\hfill
    \includegraphics[width=0.33\linewidth,height=30mm]{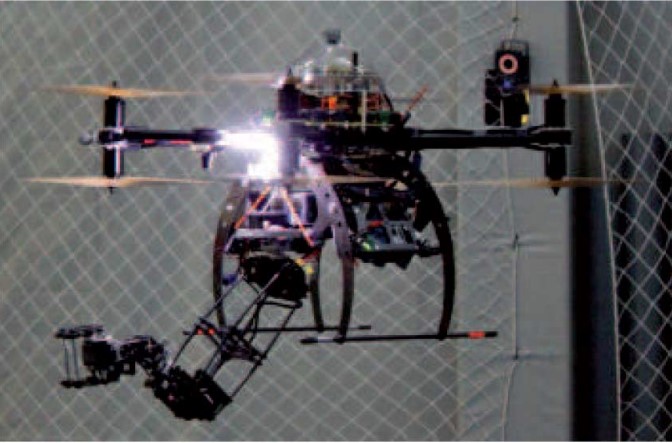}\hfill
    \includegraphics[width=0.33\linewidth,height=30mm]{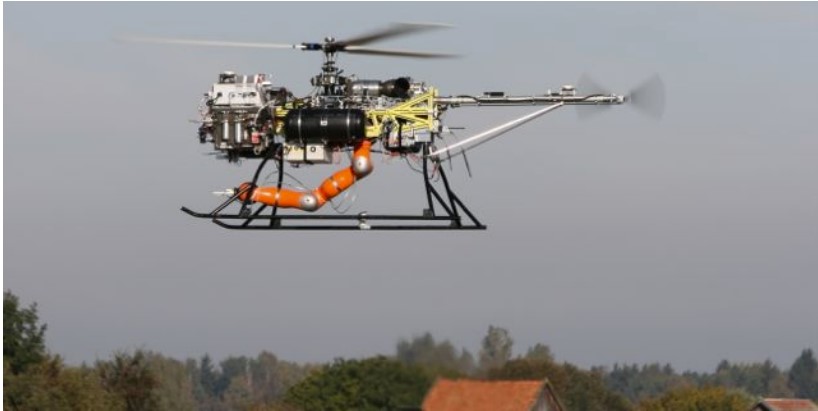}\hfill
    \vspace{0.5mm}\\
    \includegraphics[width=0.33\linewidth,height=30mm]{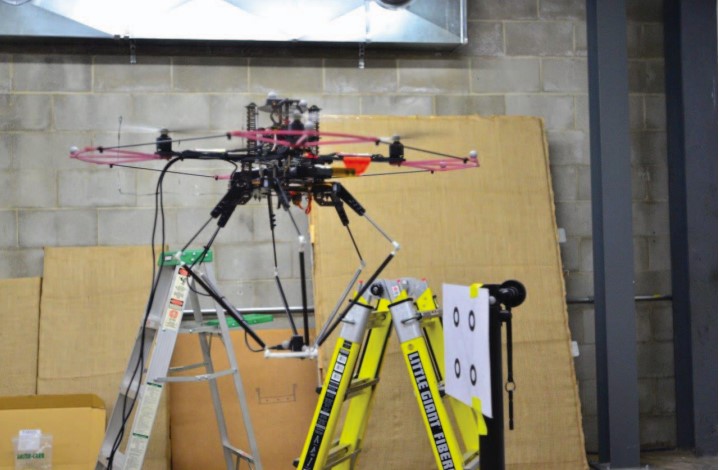}\hfill
    \includegraphics[width=0.33\linewidth,height=30mm]{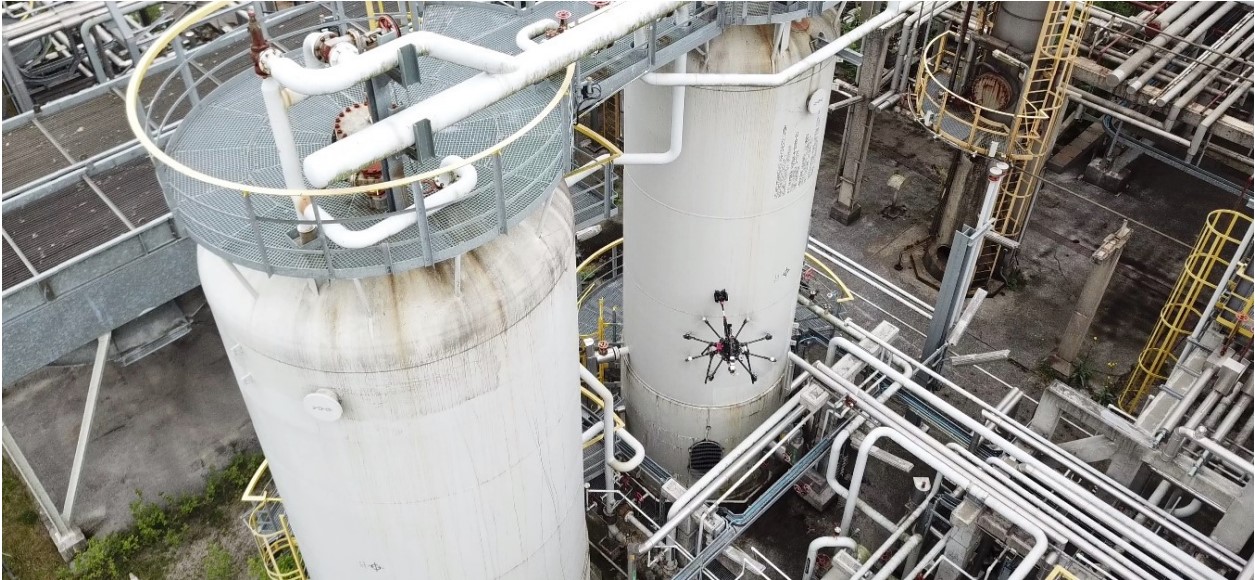}\hfill
    \includegraphics[width=0.33\linewidth,height=30mm]{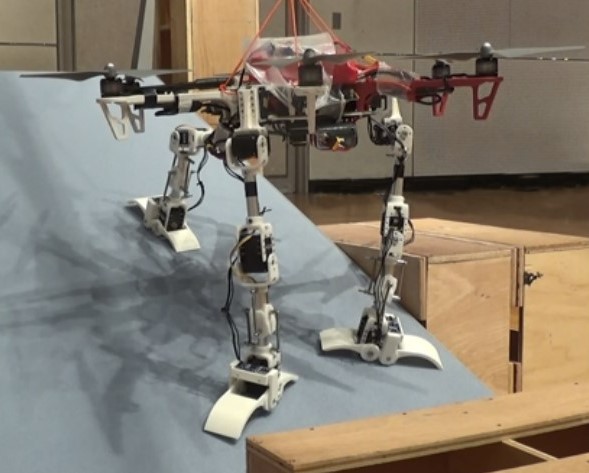}\hfill
    \vspace{0.5mm}\\
    \includegraphics[width=0.33\linewidth,height=30mm]{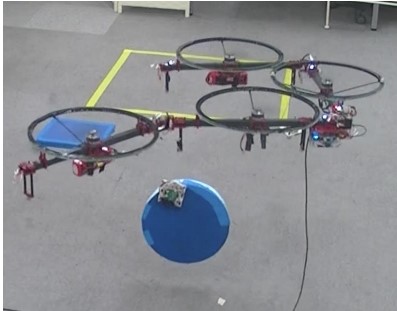}\hfill
     \includegraphics[width=0.33\linewidth,height=30mm]{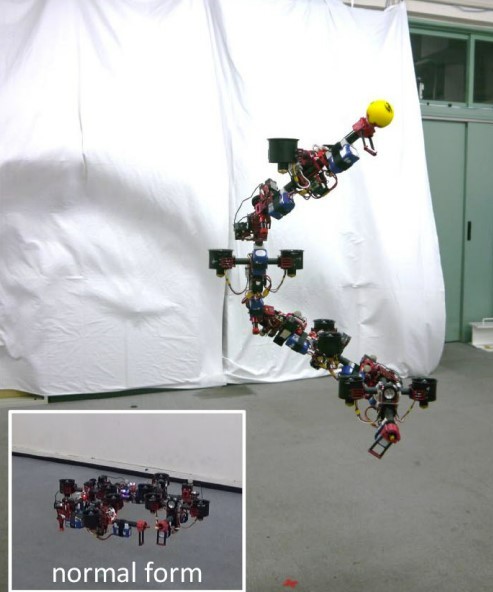}\hfill
    \includegraphics[width=0.33\linewidth,height=30mm]{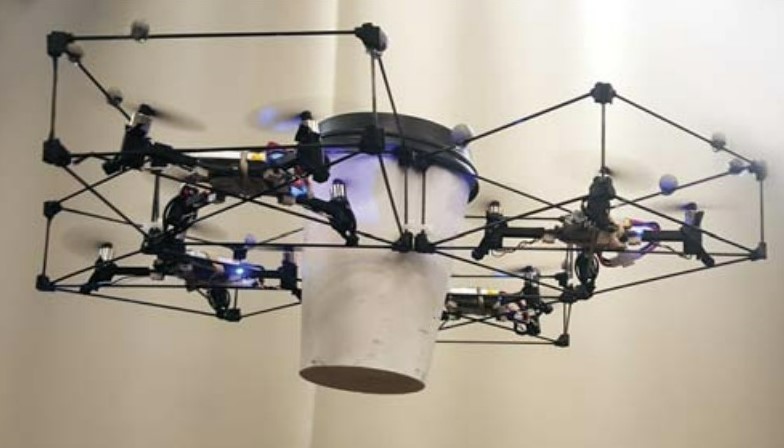}   
    \caption{Examples of Aerial Manipulation Systems: \cite{71}, \cite{77}, \cite{78}, \cite{83}, \cite{88}, \cite{90}, \cite{93}, \cite{95}, \cite{94} (from left to right, top to bottom)}\label{fig:aerial manipulation}
\end{figure}

Early research has centered on using mostly quadrotors and helicopters with embedded arms. Conducted experiments are mainly indoors using motion tracking systems, without considering perception or planning. In \cite{70}, a quadrotor using different grippers successfully grasps and transports several items. In \cite{4}, a quadrotor that is equipped with a  1-DoF arm performs assembly/construction of a cubic structure. In \cite{71}, a quadrotor applies contact forces to a wall while hovering stably, see Fig.\ref{fig:aerial manipulation}. In \cite{74}, a miniature quadrotor equipped with a 3-DoF delta structure arm and a 3-DoF end-effector successfully flew either freely or physically interacting with the environment. Other more innovative platforms include a ducted-fan miniature platform \cite{72} and a 6-DoF gantry crane equipped with two 4-DoF manipulators \cite{73}. Helicopters have also been used for load transportation using tethers \cite{30} in search and rescue applications. Grasping objects in flight, using a manipulator-gripper, is demonstrated in \cite{75}. 

Use of multirotors, like hexacopters and octacopters, which have better capabilities in precision, planning and perception, is indicative of the next arsenal for outdoors applications \cite{81}. The attached manipulator(s) have 6 and/or 7 DoF \cite{79}\cite{78} Fig.\ref{fig:aerial manipulation}, while the overall system includes onboard GNSS navigation systems, vision based controllers, hyper redundant arms \cite{80} and other robust controllers. In \cite{76}, results demonstrated successful applicability of a 4-DoF planar robotic arm equipped with a stereo camera, attached to a hexarotor. In \cite{77} a hybrid system using visual servoing and position-based control was introduced Fig.\ref{fig:aerial manipulation}. In \cite{83}, a 6-DoF parallel manipulator attached to a quadrotor was designed that demonstrated precise end-effector position Fig.\ref{fig:aerial manipulation}, while in \cite{82} authors introduced a hexarotor with a large workspace parallel 3-DoF aerial manipulator. 

Current trends in aerial manipulation include utilization of SLAM algorithms to navigate in complex environments with more accurate pose estimation \cite{86}\cite{87} and target localization \cite{103}. Such systems include platforms with tilted or tiltable \cite{104} propellers, allowing for (under actuated) multirotors to exhibit full actuation properties. In \cite{84}, a tiltable hexarotor with an attached end-effector performs inspection and maintenance tasks sliding along a horizontal plane or performing peg-in-hole tasks. In \cite{85}, a tilted hexarotor with a 2-DoF lightweight arm performs push-and-slide tasks on curved surfaces. Experiments in a refinery with realistic scenarios for pipe inspection are demonstrated in \cite{88}; a tiltable octarotor equipped with a 6-DoF robotic arm and an end-effector with wheels and embedded inspection sensors performs semi-autonomous point contact and slide tasks, see Fig.\ref{fig:aerial manipulation}. 

Several prototype systems address perception without using markers. In \cite{91}, an uncalibrated image-based visual servo strategy is proposed to drive the arm end-effector to a desired position. In \cite{91}, Random Tree Star (RRT*) algorithms are implemented to conduct outdoor path planning experiments of a multirotor with two arms for long-reach manipulation in cluttered environments, addressing multiple arm utilization. In \cite{90}, a quadrotor equipped with three 3-DoF arms demonstrates landing in uneven terrains Fig.\ref{fig:aerial manipulation}. In \cite{89}, 4-DoF anthropomorphic dual arms are integrated with a hexarotor, while in \cite{102}, a dual arm manipulator in a cable suspended configuration is proposed for aerial tool delivery to human operators for power line inspection.

Aerial manipulation may also be classified in categories that depend on the nature of the manipulation task. These categories reflect cooperative aerial manipulation (that is reviewed in this paper), teleoperation, and interconnected actuated multibody designs. The latter is a newly introduced class of aerial manipulation with mechanically connected actuators capable of changing their shape while flying. In \cite{93}, a transformable configuration that consists of four rotors separated by 4-links is introduced, see Fig.\ref{fig:aerial manipulation}. Between the links, a servo-motor allows for the system to move in the horizontal plane and, therefore, to change its shape. In conducted experiments the platform is capable of carrying two cable hanging objects while keeping its balance and stability. A similar platform with a flying gripper system is proposed in \cite{94}; four modular robots are surrounded by a carbon fiber cage, hence, attached to each other with magnets, see Fig.\ref{fig:aerial manipulation}. They create an aperture in the middle of the attachment, the docking mechanism of the payload. In \cite{95}, authors propose a fully-actuated serial-link structure platform that can change its shape while flying, and at the same perform object manipulation, see Fig.\ref{fig:aerial manipulation}. In \cite{96}, a large scale skeleton with distributed rotor actuation may perform valve tuning tasks while tethered to the ground. 

For teleoperation, regulatory and safety needs require more human intervention and supervision. In \cite{97}, a teleoperation framework for psychical contact with the environment is introduced. A human operator utilizes a haptic device to navigate a quadrotor through a virtual environment and to apply forces on surfaces with a rigidly attached passive tool. In \cite{98}, a swarm of quadrotors, with a fixed tool attached, may transport objects in a virtual environment by getting commands from motions of a human hand; fingertip motions are tracked using an RGB-D camera. In \cite{99}, indoor and outdoor experiments are conducted that demonstrate different manipulation tasks. A flying robot (SAM \cite{100}) using onboard visual sensors helps the user navigate the end-effector in order to grasp an object and through virtual reality achieve 3D visual feedback. 

\section{Cooperative Aerial Manipulation} 
\subsection{Cable-Driven}

\begin{figure}[thpb]
    \includegraphics[width=0.33\linewidth,height=30mm]{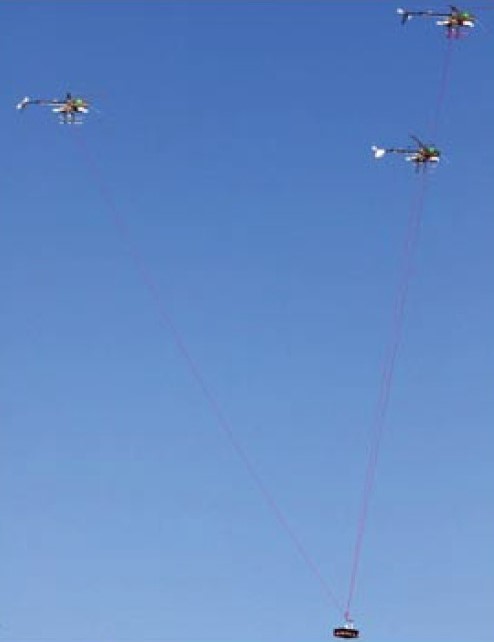}\hfill
    \includegraphics[width=0.33\linewidth,height=30mm]{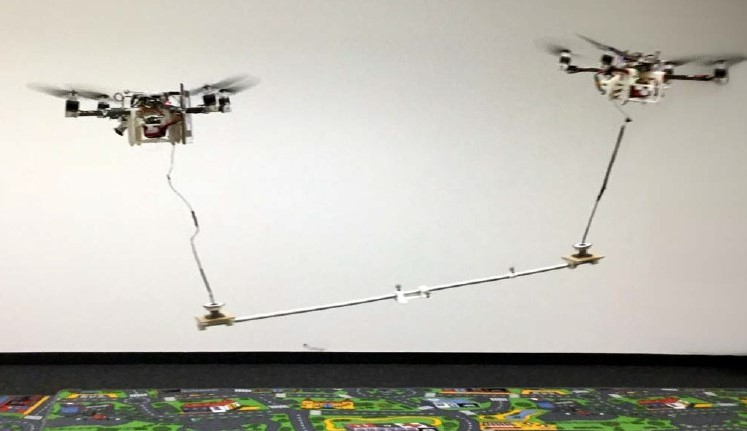}\hfill
    \includegraphics[width=0.33\linewidth,height=30mm]{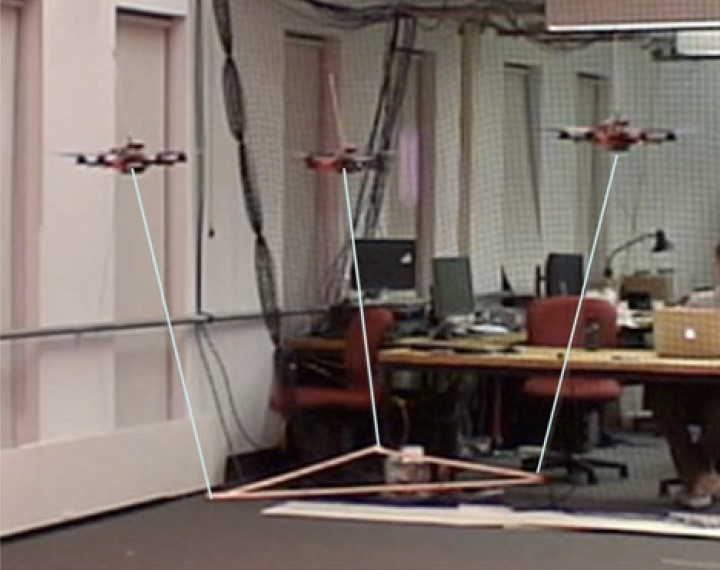}\hfill
    \vspace{0.5mm}\\
    \includegraphics[width=0.33\linewidth,height=30mm]{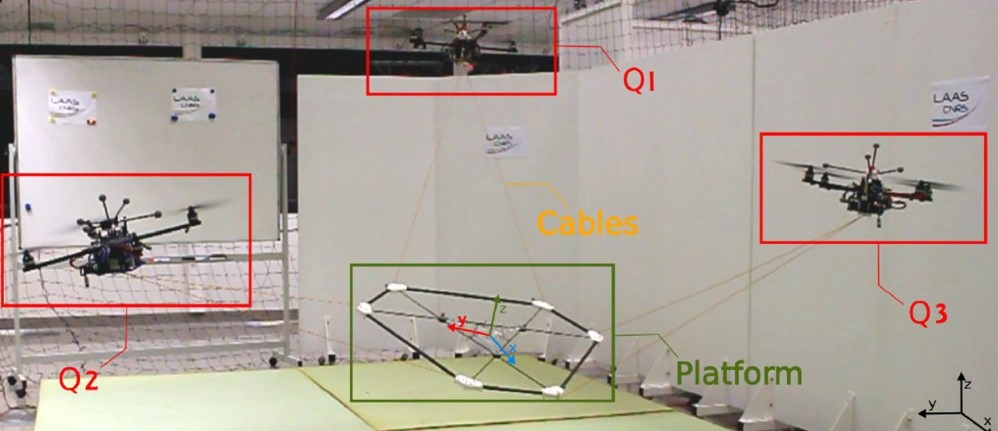}\hfill
    \includegraphics[width=0.33\linewidth,height=30mm]{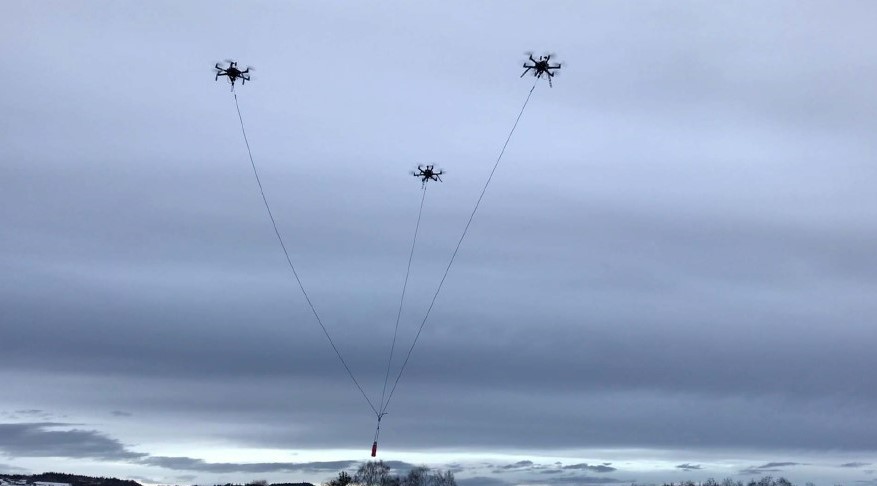}\hfill
    \includegraphics[width=0.33\linewidth,height=30mm]{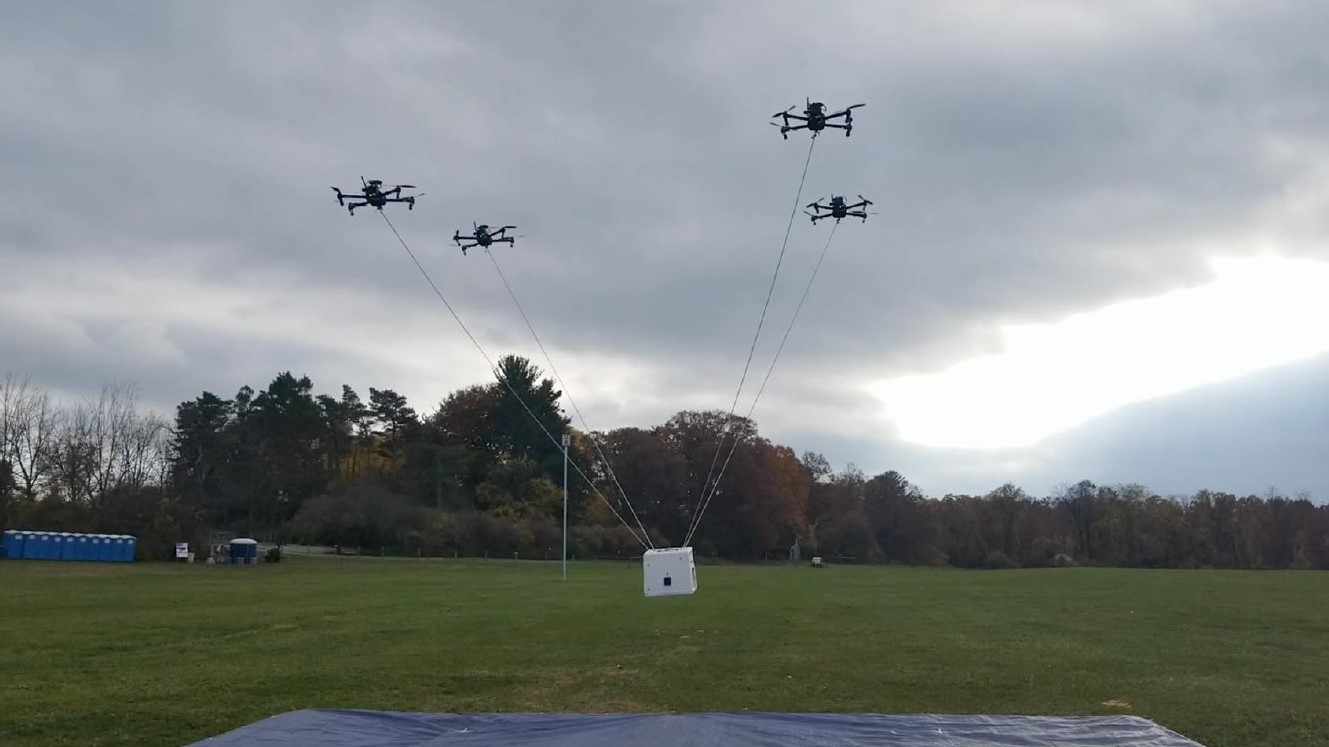}\hfill
    \vspace{0.5mm}\\
    \includegraphics[width=0.33\linewidth,height=30mm]{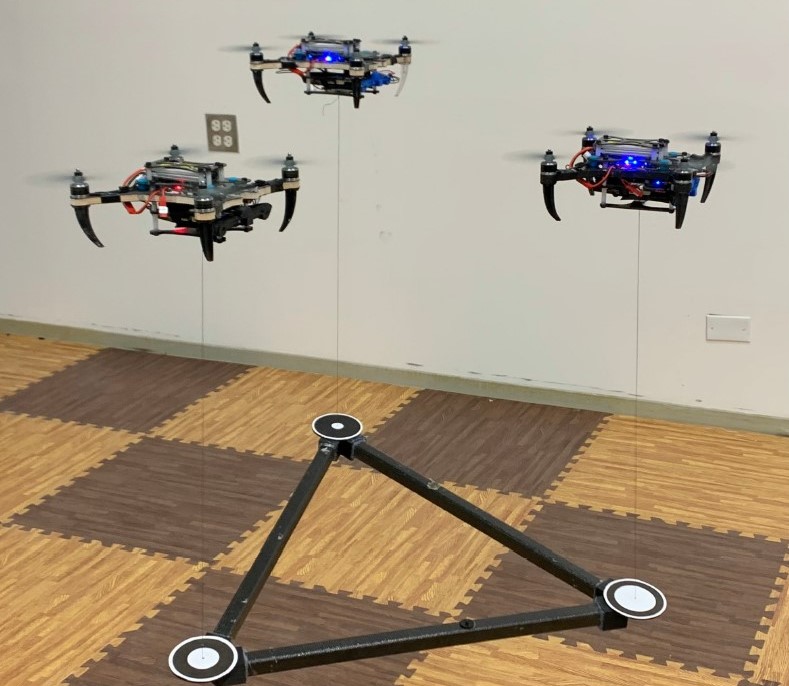}\hfill
     \includegraphics[width=0.33\linewidth,height=30mm]{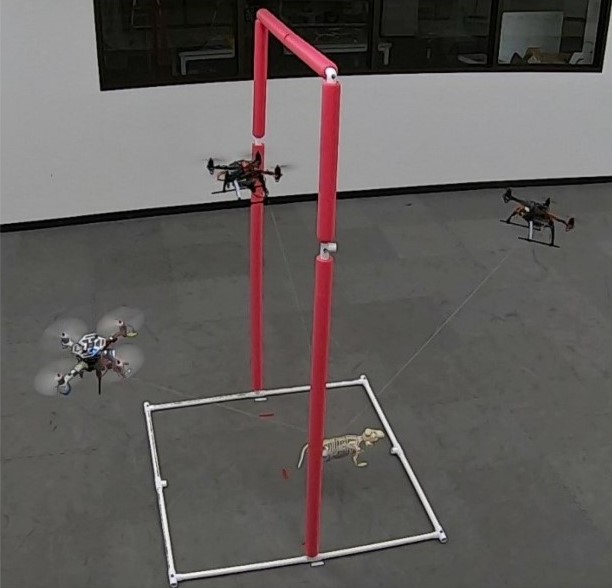}\hfill
    \includegraphics[width=0.33\linewidth,height=30mm]{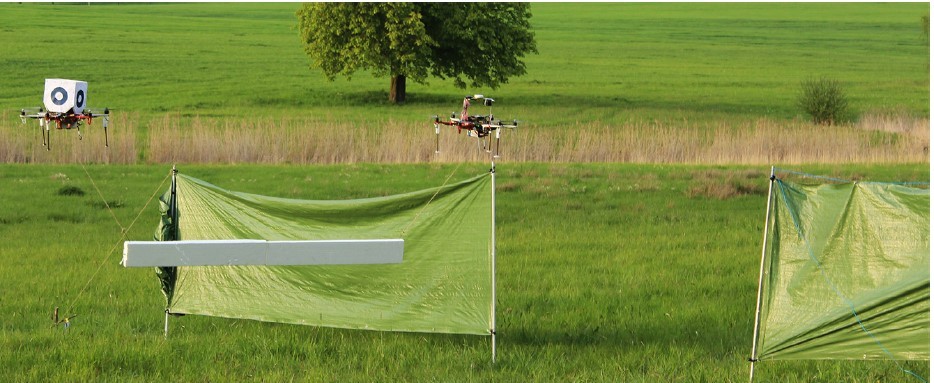}
   
    \caption{Cable-Driven Aerial Transportation: \cite{30}, \cite{31}, \cite{36}, \cite{37}, \cite{40}, \cite{57}, \cite{59}, \cite{62}, \cite{69} (from left to right, top to bottom)}\label{fig:cable-driven}
\end{figure}

The most researched area in cooperative aerial manipulation relates to cable-suspended multirotors that carry an object. In \cite{26}, three quadrotors manipulate a triangular payload connected with cables, see Fig.\ref{fig:cable-driven}. The mathematical model with forward and inverse kinematics is presented considering inter-robot collisions and cable-tension (positive tension). This approach suffers from multiple solutions in forward kinematics and from inability to reduce/eliminate oscillations resulting in slow motions, see VI. But in \cite{27}, the authors address and solve the multiple solution problem by applying cone-constraint conditions, achieving a unique payload solution. In \cite{37}, an uncertainty-aware controller for a Fly-crane system is developed and it is experimentally tested, see Fig.\ref{fig:cable-driven}.

Previously published research \cite{26}\cite{27}\cite{37} has used quasi-static models without considering the dynamics of the payload, or high-speeds. In \cite{28}, the authors introduced a differentially-flat hybrid system for cable-suspended with a point-mass load and for a rigid-body load. Simulations and experimental results with the same triangle-payload with a three quadrotors setup, have shown that the hybrid method is superior compared to quasi-static models.

Field experiments in the presence of high wind gusts are illustrated in \cite{29}\cite{30}; a slung-load transportation scheme using small helicopters is introduced with potential in carrying heavier loads, see Fig.\ref{fig:cable-driven}. A similar system has been evaluated when functioning under wind gust profiles in \cite{115}, although only in simulation. Outdoor experiments have been conducted in the presence of wind using three network-delayed quadrotors that carry a spherical load \cite{122}.

In \cite{31}, two quadrotors transport a cable-suspended load, a rod attached via magnets, at moderate speeds, see Fig.\ref{fig:cable-driven}. This is achieved using only on-board visual sensing and without any explicit communication. The control is based on the leader-follower approach and to reduce complexity, the authors consider trajectory in a straight line at a constant height. On-board state estimation has been adopted in \cite{117}. Similar work using leader-follower control and transporting a rod appears in \cite{32}; authors propose a decentralized leader-follower control with communication based on sensing of contact forces. In \cite{57}, the leader is the payload (i.e., foam box with an on-board computer), and rather than treating the latter as a disturbance, it is explicitly controlled, see Fig.\ref{fig:cable-driven}. Successful results are demonstrated in indoor and outdoor environments. After extensive experimental demonstration with two or three quadrotors and different payload weights, authors in \cite{61} mention that transportation with no consideration of payload motion or disturbances, and with no measurements of relative positioning, is feasible in outdoor environments. Similar to \cite{57} and \cite{109}, payloads have a hardware configuration; they are customized to accommodate Ethernet cables for communication between vehicles. In \cite{59}, vision-based state estimation, see Fig.\ref{fig:cable-driven}, demonstrated successful experiments with three micro air vehicles (MAVs) carrying a triangular object. A similar experiment was presented in \cite{123}. Authors proposed a nonlinear model predictive control (NMPC) method enabling manipulation in all 6 DoF. As opposed to vision-based methods \cite{59} \cite{31}, authors in \cite{119} investigate a vision-less control scheme that focuses on the follower quadrotor; the leader is controlled by an operator while the follower estimates its state from an onboard IMU based on a two-stage EKF-based estimation strategy. Authors claim that this method is suitable for small UAVs with limited computational requirements, although the payload is modelled as a point mass. Nevertheless, indoor and outdoor experiments gave produced successful results. 

An alternative approach, instead of using a master-slave setting, is adopted in \cite{40}. Instead of following the leader, UAVs are working cooperatively to achieve the same goal; outdoor experiments have been completed using three hexarotors carrying a rigid object, see Fig.\ref{fig:cable-driven}. Passivity-based control has been used in \cite{114} where three quadrotors have been used to carry a spherical object.

In contrast to traditional cable-driven parallel UAVs with fixed anchor points, authors in \cite{38} propose a more challenging approach of a wrench set tethered multi-rotor with moving pulley anchor points, providing greater flexibility. The payload maximum acceleration is evaluated by taking advantage of the newly introduced capacity margin index. Feasible trajectories are optimized based on a tension distribution algorithm. Successful experiments have been conducted in indoor and outdoor environments; for the latter all computations are done on the onboard computer.

Cooperative transportation of a non-uniform payload is addressed in \cite{39}. Authors implemented a path planning technique with RRT* algorithms combined with B-Spline curve to avoid known obstacles. Simulation results were successful, however, the system strongly depends on reliable communication. In \cite{120}, real-time formation planning in obstacle filled environments is considered without relying on payload and cable length. Dynamic formation in obstacle filled environments is presented in \cite{121} without any assumption on the cable's tension, nor the need for payload's measurements; the payload's geometry is not taken into account.

Coupled dynamics and geometric control is investigated in  \cite{41}\cite{42}\cite{43}\cite{44}\cite{113}. In \cite{41} authors propose cooperative transportation of a suspended load using quadrotors based on specific assumptions of point load mass and rigid without mass links. In \cite{43}, the point mass assumption is replaced with a rigid load connected via flexible cables where each flexible cable is modeled as a system of serially-connected links. In both \cite{41} and \cite{43} successful simulation results are included. Work in \cite{43} is extended in \cite{44}, where the theory is demonstrated with experimental results that include stabilization of the payload in the presence of external disturbances. Moreover, to solve the direct relation between the motion of the quadrotor and the motion of the payload, authors in \cite{58} focus on reconfigurable cable-driven parallel robots (RCDPR). This approach is more suitable for teleoparation, though. It has been validated through physical simulations with a human-in-the loop present. In \cite{113} authors propose a method that has advantages over the previously reported research in \cite{41} and \cite{42}. The geometric nonlinear controller does not require the linear and angular acceleration of the payload as well as the link information; thus, it is easier to implement in a real life scenario.

As opposed to previous research on cable-driven manipulation, authors in \cite{60} do not assume that the cables are connected to the object. Instead, they automate the whole process by developing a system composed by two quadrotors connected by a hanging cable. The quadrotors wrap the cables around the cube-shaped object and pull. The box has no hooks; manipulation  depends only on the friction between the box and the cables. This method has potential in fully autonomous operations, although demonstrated only in simulations. 

In \cite{62}, the focus is on simulating real world conditions in outdoor environments (different weather conditions) in the presence of obstacles. Two hexacopters transport a large object through narrow passages, see Fig.\ref{fig:cable-driven}. Sampled-based path planning is developed to maintain the desired distance between the two UAVs during transportation, to prevent oscillations and increase robustness. 

Most recent research centers around heavy payloads \cite{68}\cite{69}, high speeds \cite{67} and robotic arms with cables \cite{108}. In \cite{68} four multirotors transport a tethered heavy object under motion formation control. Implemented algorithms only require knowledge of the relative position measurements from the robots with respect to their neighbours; This approach is suitable for GPS-denied or adverse environments. In \cite{69} three multirotors transport a heavy object while passing though a narrow doorway, see Fig.\ref{fig:cable-driven}. The algorithm is efficient and runs on resource-constrained onboard computers. In \cite{67}, two multirotors transport a rod with speeds up to 1.6 m/\(s^2\) using adaptive control. In \cite{106} anti-swing control is shown for a team of quadrotors performing cargo delivery tasks; the method showed enhanced performance in cargo swing suppression. An alternative approach for payload tracking, instead of utilizing markers as in \cite{57} and\cite{59}, is presented in \cite{108}. A system of two aerial manipulators is adopted with a slung load connected to the end-effectors. The aerial manipulator tracks the load position. Although successful simulation results have been produced, the system has not been tested experimentally. Considering combined cable-drive systems (e.g. \cite{108} manipulators with cables), authors in \cite{116} developed a bar spherical joint structure that includes four quadrotors with four attached cables. This design increases safety during flight; according to the authors this design reduces energy consumption. In a conduced lab experiment it was demonstrated that the system could carry a heavier payload than common cable-suspended systems. In \cite{118} the underlying leader-follower system has been studied as as a  nonholonomic one; the leader tracks the trajectory by receiving data from an IMU connected to the payload; the follower is not allowed to undergo transverse motion. This method has no inter-agent communication. Lab experiments have shown proof-of-concept demonstration feasibility. However, authors do mention that tracking errors seem to be affected from airflow and that a more-practical IMU setup should be considered. In \cite{124} authors mention that even drones of the same type share inconsistencies, i.e., thrust uncertainties, aerodynamic and hardware uncertainties, which need to be modelled and controlled to get accurate cable force estimates. During experiments a heterogeneous system is used with two different weight quadcopters that are connected to a pipe by two uneven cables; through force-consensus control the pipe tries to align with the ground, achieving shared mass distribution. Obtained results showed successful demonstration under quasi static conditions.

\subsection{Aerial Manipulation with multi-DoF arms}

Tether-based approaches are popular since the tether that is connected at the center of mass of a multirotor does not generate additional torques on the multirotor \cite{12}. Although this method supports static equilibrium of the robot and object, the object cannot be directly adjusted or manipulated. This limitation is overcome by combining aerial robots with multi-DOF robotic arms, functioning as one system. 

Authors in \cite{10} solve the complicated dynamics of multiple quadrotor-manipulator systems and compute the cooperative force distribution among the robotic arms by proposing a hierarchical control framework, see VI. The proposed framework is versatile and adaptable as it allows the modification of only one block instead of redesigning the controller. Nevertheless, to bypass the dynamics complexity, authors in \cite{19} propose two impedance control laws to limit external and internal forces and to track the end effector's reference trajectory with the rigid object grip assumption. However, results in \cite{10}\cite{19} are validated only via simulations. 

\begin{figure}[thpb]
\includegraphics[width=0.33\linewidth,height=29.8mm]{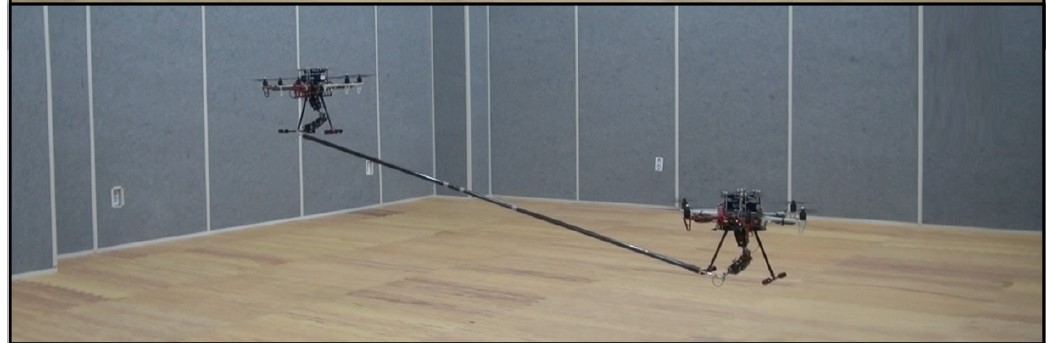}\hfill
\includegraphics[width=0.33\linewidth, height=30mm]{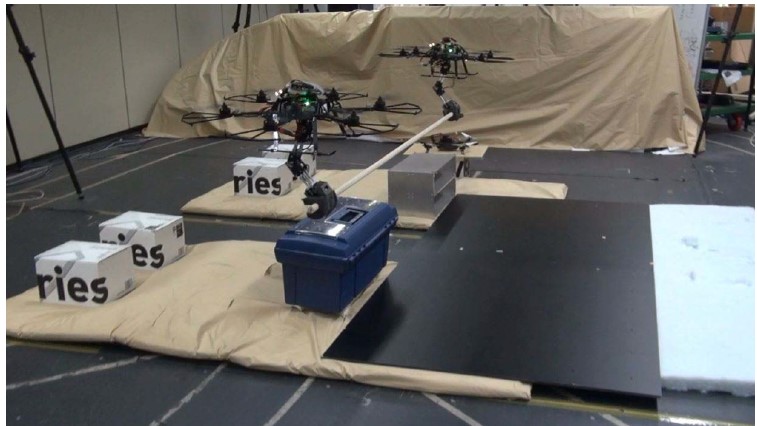}\hfill
\includegraphics[width=0.33\linewidth,height=30mm]{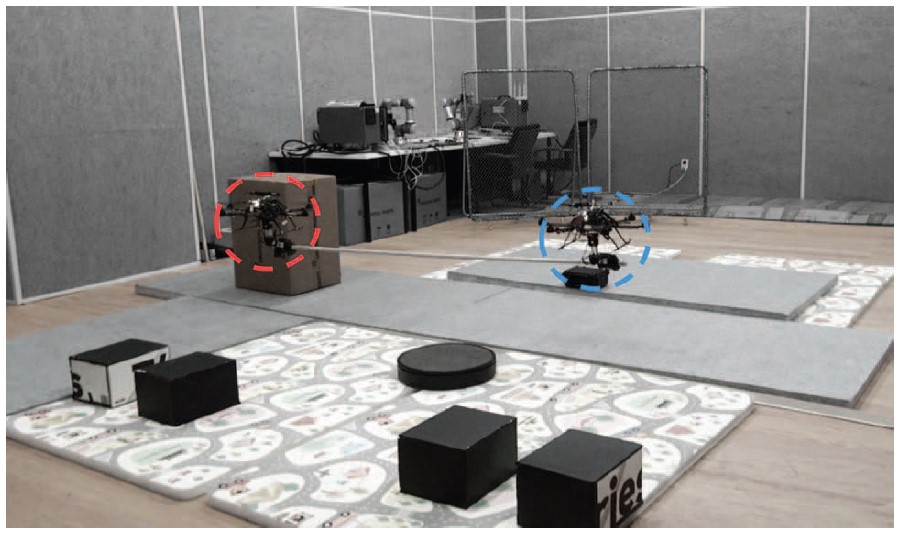}
\caption{Cooperative Aerial Manipulation of a Rod With Multi-DOF Robotic Arms: \cite{12}, \cite{16}, \cite{20} (from left to right)}\label{fig:robotic arms}
\end{figure}

Most experimental work in this category is lab-based. Two hexarotors equipped with multi-DoF arms (3-DoF \cite{12}, 2-DoF \cite{15}) manipulate the pose of a rod-shaped object, see Fig.\ref{fig:robotic arms}. A first demonstration is presented in \cite{17} where authors propose an augmented adaptive sliding mode controller that accounts for the object's dynamics, considering constraints about the grasping point. The desired path for each aerial manipulator is obtained by using an RRT* algorithm to transport the object to the desired position. Results in \cite{17} are extended in \cite{16}, as the path planning algorithm (RRT*) is enhanced with Bezier curves to produce the required trajectory in known environments; dynamic movement primitives (DMPs) are utilized to avoid unknown obstacles. As for grasping, a commercial gripper is used in experiments, see Fig.\ref{fig:robotic arms} instead of the handle/hook configuration adopted in \cite{17}. Aiming at real-time implementation, results are further extended in \cite{20} where a learning-based motion planner is proposed using a combination of RRT* algorithm and parametric-DMPs (PDMPs). This method achieves faster and more efficient manipulation with concurrent object avoidance in an unknown environment, see Fig.\ref{fig:robotic arms}. Authors mention that the trade-off between motion optimality and computational time of the proposed algorithm is balanced effectively in comparison to conventional approaches (e.g. sampling-based, optimization-based). In \cite{15}, authors propose a framework for realistic applications that consists of an on-line estimator for the mass/inertial properties of an unknown payload, and an adaptive controller. Additionally, real-time obstacle avoidance is achieved by using an RGB-D camera and DMPs. Results are extended in \cite{18}, where authors consider velocity and curvature constraints for collision-free, real-time smooth path planning algorithms. The proposed motion planning optimization approach achieves reduced computational costs. Although in \cite{18}, \cite{15}, \cite{20} Vicon systems still measure position information, re-planning is performed onboard. A successful demonstration with 3-DoF arms is presented in \cite{12}, although the end-effector is still rigidly attached to the object. By employing a robust control scheme and an NSB approach, see VI, this method succeeds in safe manipulation (i.e., collision avoidance between the rod and aerial vehicle) and internal force estimation without the use of force sensors (expensive, power consuming).

For simplicity reasons, previously published research has focused on manipulating objects with uniform mass distribution. Recent research in \cite{11} focuses on the control of an arbitrary mass object using two aerial manipulators. The control framework uses two estimation parameters; kinematic for relative distances and dynamic for the mass/ center of mass of the payload by using an external wrench estimation algorithm. The results are validated via simulations under assumptions (e.g. PE (persistence of excitation conditions, previously known configuration of the payload). In \cite{65} and \cite{66}, authors present force-consensus and force control methods for cooperative aerial manipulators, which can be applied to a broader range of applications using rigid and flexible payloads. Both demonstrated in simulations.

Recent research in \cite{112}, focuses on transportation without using any communication; a leader dictates motion (i.e., markers on the platform) and the followers use computer vision (i.e, equipped with RGB-D Cameras) in order to autonomously maintain formation with respect to the leader.

\subsection{Flexible Payload}

\begin{figure}[thpb]
    \centering
    \includegraphics[width=0.33\linewidth,height=30mm]{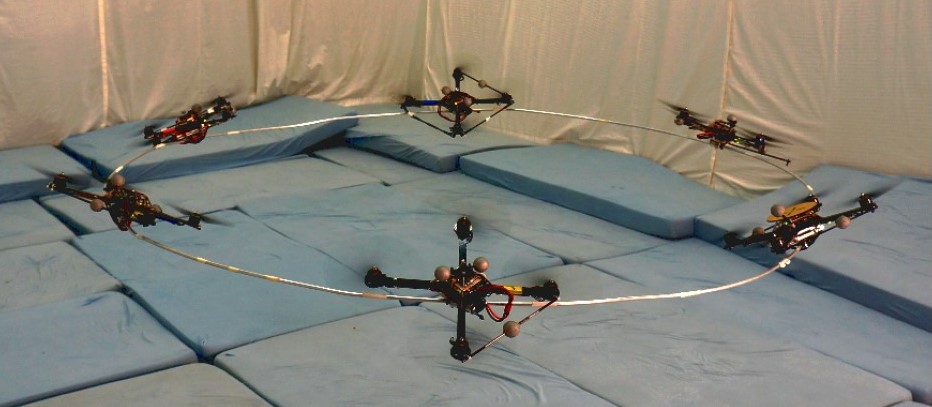}\hfill
    \includegraphics[width=0.33\linewidth,height=30mm]{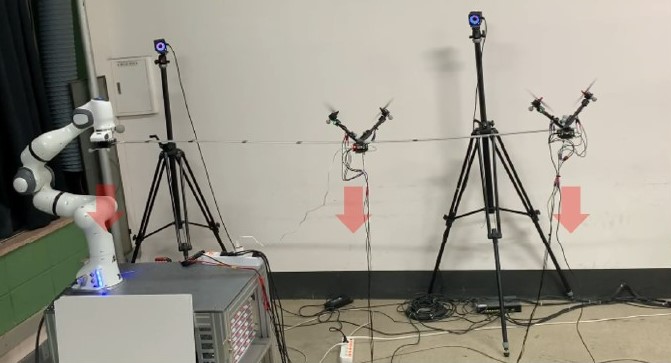}\hfill
    \includegraphics[width=0.33\linewidth,height=30mm]{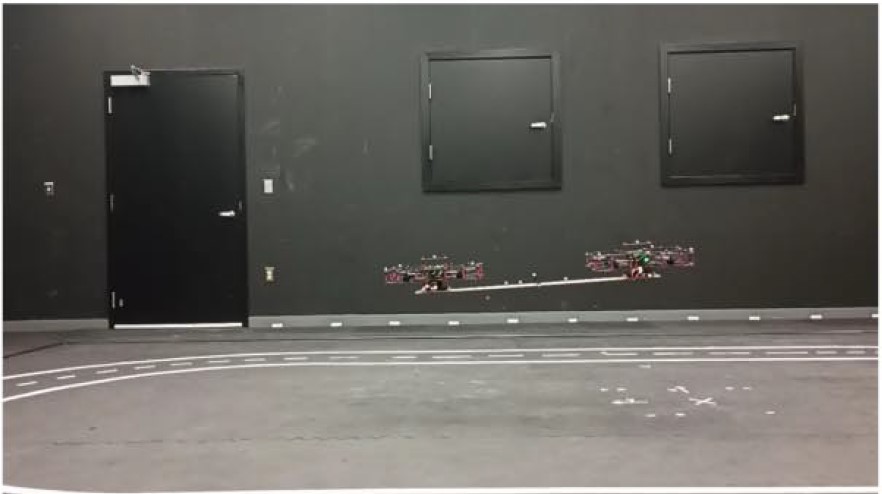}
    \caption{Cooperative Aerial Transportation of a Flexible Object: \cite{21}, \cite{22}, \cite{23} (from left to right)}\label{fig:flexible payload}
\end{figure}

Cooperative manipulation of a flexible payload has not been widely investigated. In \cite{21}, six quadrotors are rigidly attached to a thin-flexible ring, positioned with a titled angle, to perform transportation tasks in lab experiments, see Fig.\ref{fig:flexible payload}. Control design, see VI, is based on a linearized system around hover conditions that is capable of controlling deformations and desired positions. However, the system performs only horizontal transportation.

Based on \cite{51} and \cite{52}, in \cite{22}, authors propose a distributed rotor-based vibration module (RVM) for the manipulation of a 2m aluminum bar. For the RVM design, authors consider maximized thrust with minimal torque along the plane that vibration occurs. They also utilize the controllability grammian to solve the optimal placement of the RVM on the flexible load. Experiments are implemented with a stationary robot arm and two tilted-rotors and they demonstrate successful vibration suspension, see Fig.\ref{fig:flexible payload}. 

Authors in \cite{54} extend the work of \cite{51} by considering a flexible payload. The controller succeeds to suppress the vibration of a long slender load, although only in simulations.

In \cite{23}, authors present cooperative transportation of a flexible beam by two octarotors, see Fig.\ref{fig:flexible payload}. To eliminate the movement that may be produced between the robot and the payload, the latter is connected with the robots using permanent magnets. In comparison with \cite{21}\cite{24}, this paper introduces a nonlinear dynamic method for the whole system, a fictitious output for the trajectory and an augmented linearized state feedback controller, see VI. Lab experiments show satisfactory results regarding trajectory performance and beam’s dynamic deformations.

\subsection{Ground-Air}

\begin{figure}[thpb]
    \centering
    \includegraphics[width=0.33\linewidth,height=30mm]{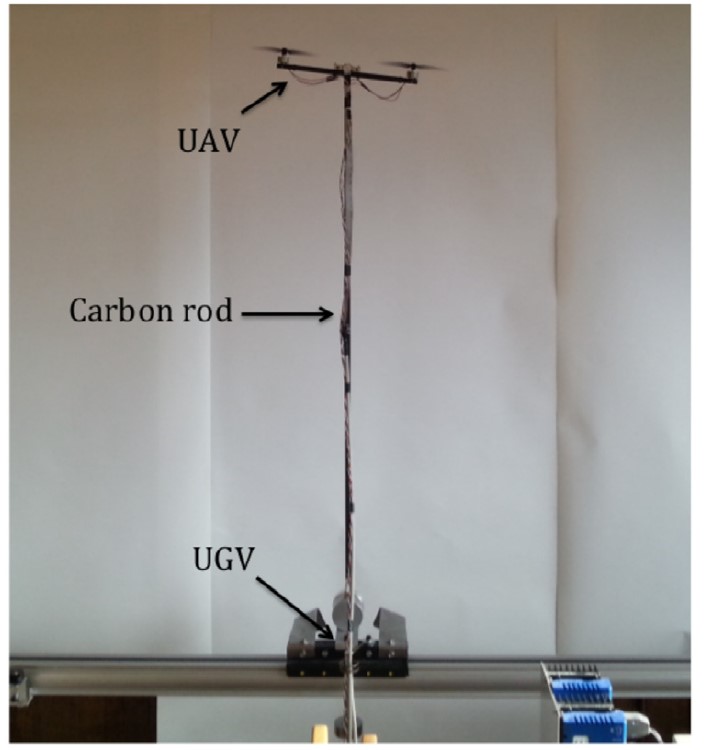}\hfill
    \includegraphics[width=0.33\linewidth,height=30mm]{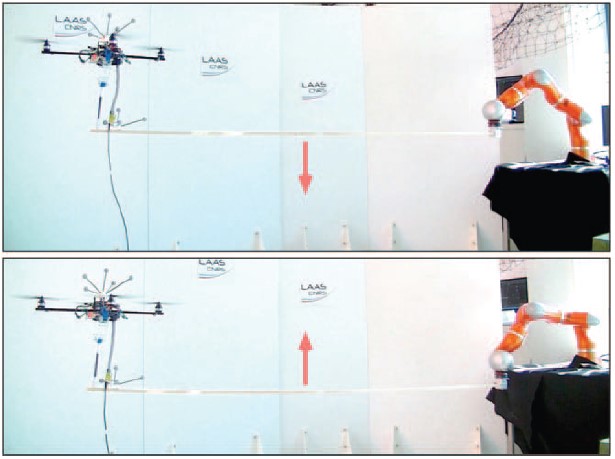}\hfill
    \includegraphics[width=0.33\linewidth,height=30.2mm]{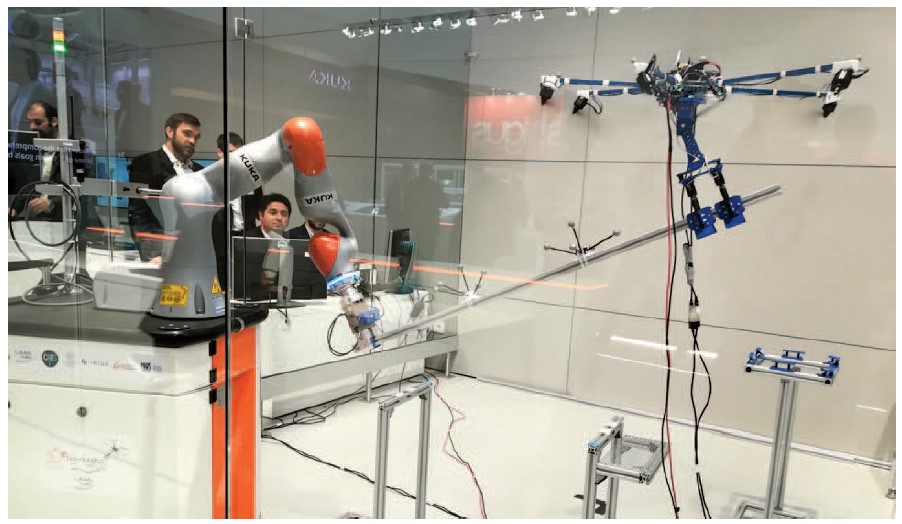}\\
    \vspace{0.7mm}
    \includegraphics[width=0.33\linewidth,height=30mm]{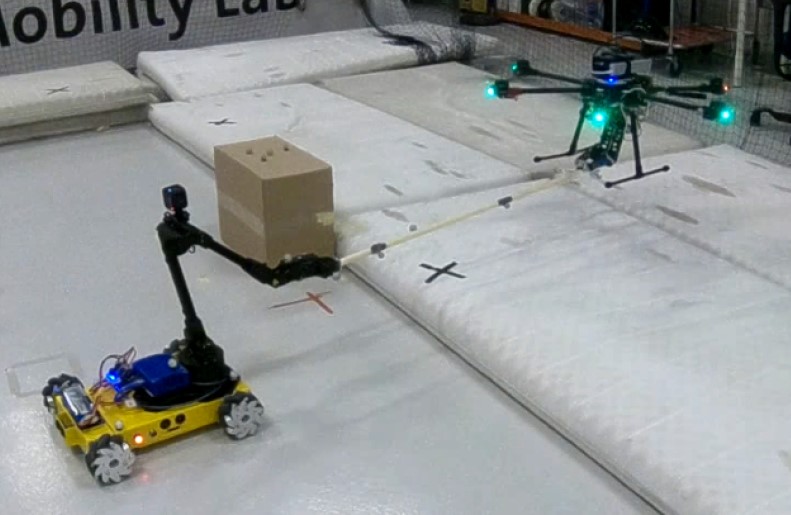}
    \includegraphics[width=0.33\linewidth,height=30mm]{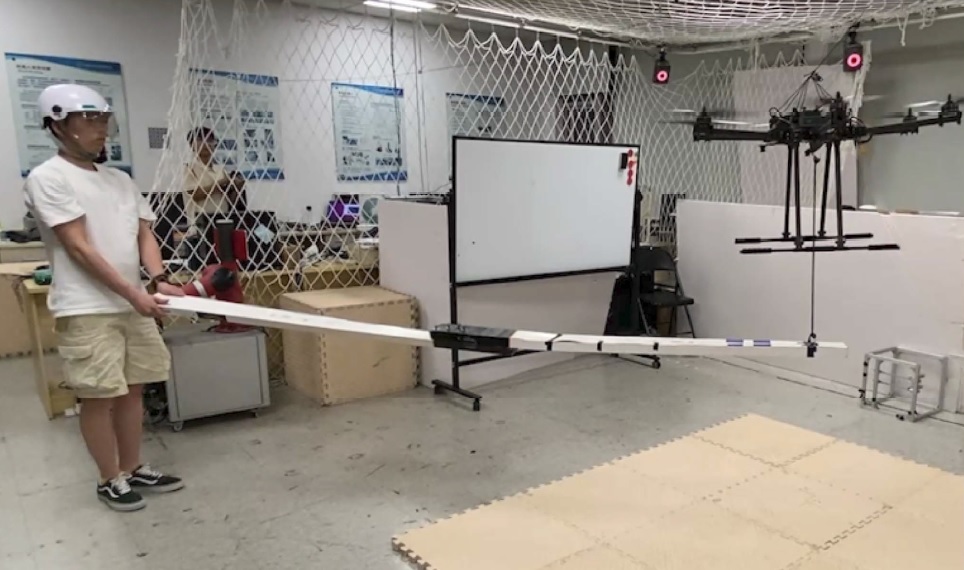}
    \caption{Ground-Air Cooperative Aerial Manipulation: \cite{50}, \cite{51}, \cite{52}, \cite{55}, \cite{110} (from left to right, top to bottom)}\label{fig:ground-air}
\end{figure}

Collaboration between Unmanned Ground Vehicles (UGV) and UAVs is a field of cooperative aerial manipulation that has emerged recently, yet beneficiary, as ground robots offer higher payload capacity while drones offer unlimited workspace \cite{14}. In \cite{50}, a quadrotor and a ground vehicle (UGV) manipulate the pose of a rod showing successful simulation and lab experimental results, see Fig.\ref{fig:ground-air}. New approaches utilizing multiple aerial-ground manipulator system (MAGMaS) \cite{51}\cite{52} offer larger workspace and movement freedom than the planar movement cart experiment presented in \cite{50}. To be specific, in \cite{51}, lab experimental results showed successful applicability of the cooperation between a ground manipulator (KUKA LWR4) and a quadrotor; both manipulate the pose of a heavy and long wooden bar, see Fig.\ref{fig:ground-air}. The ground manipulator grasps the object rigidly while the aerial robot is connected to it via a passive spherical joint. This setup was later extended in \cite{52}, where the spherical joint was replaced by the OTHex multirotor \cite{53} (i.e., tilted hexacopter with multi directional thrust ability and unique configuration allowing the manipulation of long objects), and the ground manipulator by a KUKA 7-DoF LBR-iiwa, see Fig.\ref{fig:ground-air}. Teleoperation, larger workspace and gripper actions are proposed in this work; however without yet advanced perception. Nevertheless, indoor demonstration experiments with both systems to cooperatively grasp and manipulate a rod showed successful results.

Focusing on larger workspace and enhanced capabilities, in \cite{55}, a leader-follower model predictive control (MPC) is proposed with obstacle avoidance. The setup is composed of a hexarotor and a ground vehicle; both equipped with robotic arms and grippers, see Fig.\ref{fig:ground-air}. In the lab experiments, the heterogeneous robots are holding and manipulating a plastic bar while performing maneuvers in order to avoid two different obstacles. Regarding the results, authors mention that an error is present due to imperfect low-level control tracking but the system still reacts successful to disturbances. The method relaxes from strong assumptions widely found in literature (i.e., rigid grasping robot-object points). However, the MPCs run on a off-board computer in the same network and the robots are grasping the object a priori.

Recent research focuses on human-robot collaboration for transporting objects \cite{110}\cite{111}. In \cite{111}, the human and the quadrotor, in unison, lift and transport a rod in an outdoor experiment. The operator initiates the task by grabbing the payload from one end with the help of a Human Handle Device (HHD) - a custom built sensor that for estimates human commands. At the other end, the UAV is equipped with a Cable Attitude Device (CAD), which also provides state feedback. In a similar lab experiment in \cite{110}, a vision-based velocity observer tracks human estimated velocity. This technique eliminates the need to measure the robot velocity measurement, see Fig.\ref{fig:ground-air}.

\subsection{Rigidly-Attached}

An other line of research of cooperative aerial manipulation is rigidly attached multirotors transporting a payload. This configuration can benefit from its mechanical simplicity while carrying heavier objects.

\begin{figure}[thpb]
    \centering
    \includegraphics[width=0.33\linewidth,height=30.3mm]{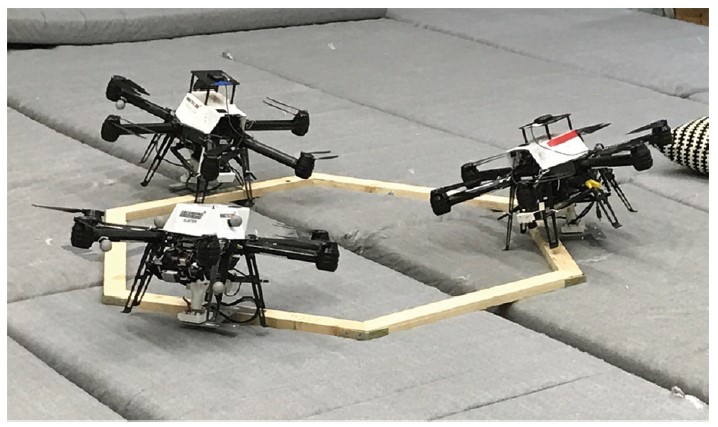}\hfill
    \includegraphics[width=0.33\linewidth,height=30mm]{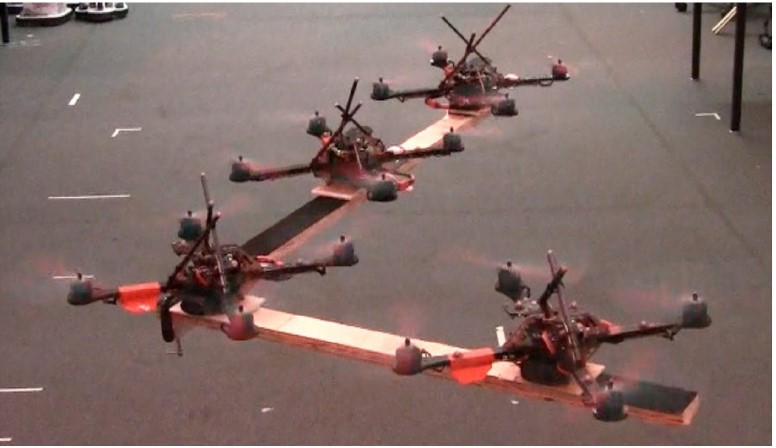}\hfill
    \includegraphics[width=0.33\linewidth,height=30.3mm]{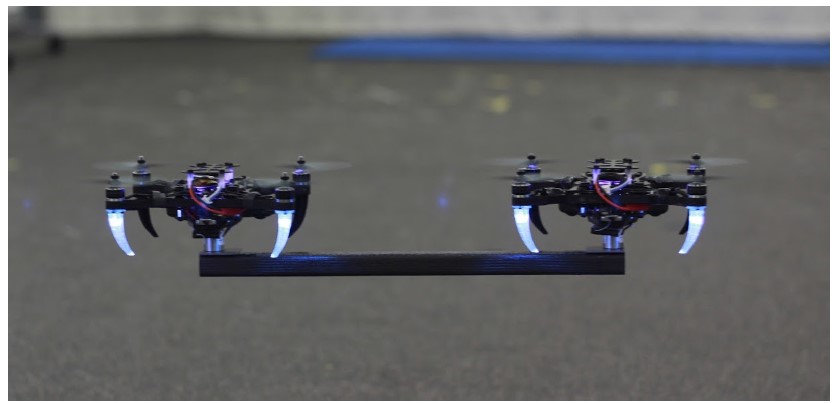}\hfill
    \vspace{0.5mm}\\
    \includegraphics[width=0.33\linewidth,height=30mm]{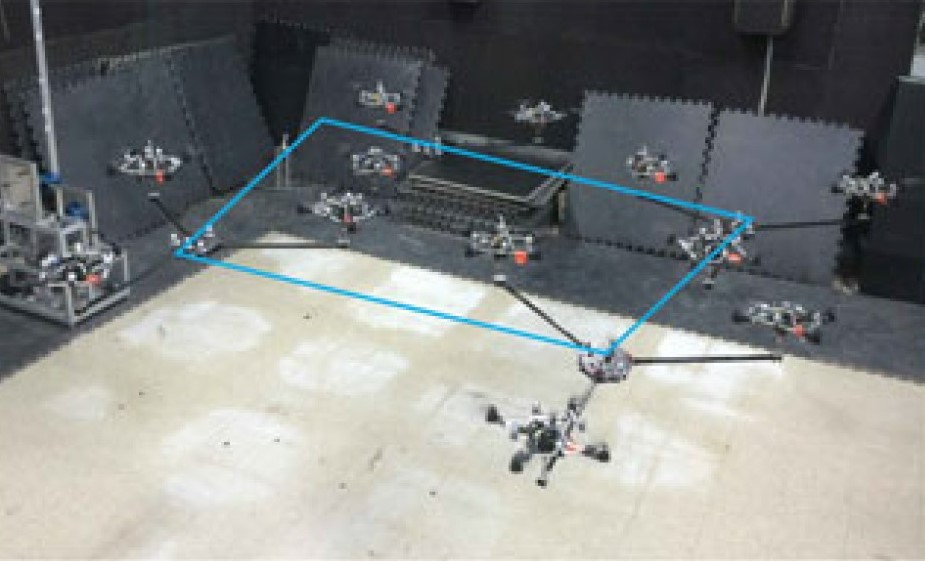}
     \includegraphics[width=0.33\linewidth,height=30mm]{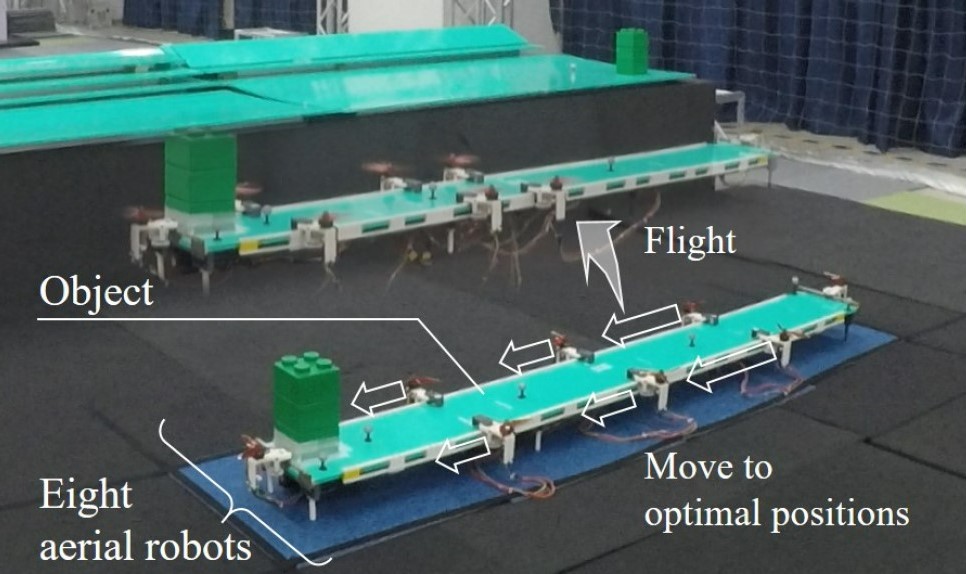}
    \caption{Cooperative Aerial Transportation with Rigid Connections: \cite{34}, \cite{45}, \cite{46}, \cite{47}, \cite{64} (from left to right)}\label{fig:rigidly attached}
\end{figure}

A first demonstration of this idea was presented in \cite{45}, where authors address the problem of cooperative grasping/transporting using multiple quadrotors. The control was implemented by combining decentralized (angular velocity) and centralized (position, velocity, orientation) formulation. The grasping mechanism consists of a gripper capable of penetrating into surfaces using microspines in order to attach/release the payload (i.e., wood), see Fig.\ref{fig:rigidly attached}. Two lab-based experiments are presented for validation; the first one was held with four quadrotors rigidly attached to a wooden planar payload with different quadrotor configurations (line, cross, L-shape, T-shape) and a second one with two quadrotors cooperatively grasping and transporting a rod. The work of \cite{45} later extended in \cite{46} where authors developed a more passive grasping mechanism for the payload, using permanent electromagnets. Based on a nonlinear controller, localization and state estimation was succeeded through onboard IMU and camera instead of external motion capture system as in \cite{45}. Lab-based experimental results with two quadrotors transporting a rod showed increased agility, see Fig.\ref{fig:rigidly attached}. Different approaches adopted mono-rotors rigidly attached to a rectangular heavy object as in \cite{63} and \cite{64}. In \cite{64}, eight mono-rotors are attached to the payload using a variable attachment mechanism (i.e., rail mechanism), allowing the robots to slide and, hence, change the optimal attachment positions depending on the payload's weight, see Fig.\ref{fig:rigidly attached}. Conducted experiments of \cite{63}\cite{64} demonstrated indoors in lab experiments. 

The results of \cite{45}\cite{46}\cite{56}\cite{63}\cite{64} are more for transportation tasks than manipulation as the payload cannot be directly controlled due to the under-actuation of the quadrotors \cite{14}. Authors of \cite{48} and \cite{47}, propose a platform that consists of multiple quadrotors connected to a frame by spherical joints (spherically-connected multiquadrotor or SmQ). The innovation of this method surpasses the limitation of under-actuation while also providing less energy consumption caused by the extra actuators. For the model and control design of \cite{47}, authors considered constraints in spherical joints range limit. The idea was demonstrated through various scenarios (i.e., motion tracking, hovering while acting force, telemanipulation), see Fig.\ref{fig:rigidly attached}. However, not much information yet exist about the feasibility of this novel approach.

Spherical joints technology was also adopted in \cite{34} for the collaborative transportation of a hexagon wooden structure, rigidly attached to three quadrotors, see Fig.\ref{fig:rigidly attached}. The system succeed to perform outdoor experiments using onboard sensing and no communication between partners. Furthermore, in \cite{49}, authors present transportation tasks by three quadrotors, connected via spherical joints to a passive adhesive tool; the method is demonstrated only in simulation with human hardware-in-the-loop.

\section{Modeling Approaches}

Modeling approaches have been widely researched \cite{5}\cite{126}. Modeling in aerial manipulation is mostly based on using either Recursive Newton-Euler (N-E) or Euler-Lagrange (E-L) formulations. The former considers the multirotor/quadcopter and the manipulator as two distinct subsystems; it is best suited for actual implementation and testing. The latter considers the multirotor/quadcopter and the associated manipulator as a single system; it is most suitable for analytical techniques and to study dynamic properties of the underlying system. In both cases, interdependence and interaction forces among the system components must be considered. Regardless of formulation, when it comes to cooperative aerial manipulation modeling complexity is increased. Next, the modeling formulations of cooperative aerial manipulation are presented, compared and summarized in Table \ref{tab:my-table2}. 

\subsection{Newton-Euler Modeling Formulation}

Considering the cable driven category, in \cite{27} the quadrotors are modeled as a team of three point-model robots; they are considered as a kinematic system while the payload is considered as a single (dynamic) rigid body. In \cite{28}, differential flatness using the N-E equations is demonstrated under two scenarios; point-mass load and rigid-model load. Then, the individual models are enhanced considering that the tension in any of the cables drops to zero - this results in a hybrid system model. A point-mass load is also considered in \cite{67} and \cite{69}. The N-E equations are modified to use quaternions that are more suitable for aggressive maneuvers and large angular displacements. In \cite{32}, aerial robots and a beam-like payload are modeled based on the N-E formulation but also considering the tension of the cable. A double integrator is developed for the closed loop translational dynamics. This method is applicable to multidirectional and unidirectional platforms. Research in \cite{28}, \cite{69} \cite{123} and \cite{32} assumes taut cables for modelling purposes, while in \cite{27} the cables are considered slack. In \cite{57}, aerodynamic effects and different cable models (like inextensible cable, spring/damper cable model, and rigid formation) are derived. In \cite{110} a cable is modeled as a unilateral spring with negligible mass. In \cite{109}, where no assumption is made about the cable's state, the N-E method describes the dynamics of the system, while graph theory (e.g., incidence matrix) is followed to indicate the communication between quadrotors and the payload. In \cite{114}, the cooperative transportation system is modelled without restricting the attachment point of the cable on the drone. In \cite{122}, the quadrotors-load modelling approach considers the input delay in the equations of motion. In \cite{124}, the system dynamics consider the thrust uncertainty, while the cable force is treated as a lumped disturbance.

In \cite{66}, the dynamic model of the Unmanned Aerial Manipulator (UAM) system is derived following the recursive N-E method. The dynamics of the whole system (i.e., UAMs carrying an object) are derived by utilizing a leader/follower formation; the carried object acts as the 'coupler' between the pair of UAMs and the exerted forces from both robots' grippers are related to the object's stiffness. The same strategy is adopted in \cite{112} considering the wrench vector measured at the end-effectors' force-torque sensor. The N-E approach is also used in \cite{65}, considering the deformation of the payload due to contact forces. In \cite{21}, a system of N quadrotors attached to a flexible payload is also modeled using a N-E formation that accounts the payload deformation.  

In \cite{45}, a system of N quadrotors is modeled in which a payload is attached. The formulation allows for quadrotors to be mounted in different planes. The same equations of motion were also used in \cite{63}, \cite{46} and \cite{56}. In \cite{63}, the equations of motion were further extended to consider robot failures in the model. However, in \cite{56}, as opposed to \cite{63} and \cite{46}, quadrotors are not placed in a parallel plane. In \cite{47} and \cite{48}, the N-E dynamics of N quadrotors and SmQ frame with the assumption that the spherical joint is attached at the CoM of each quadrotor; spherical joints limits are also included in the model. In \cite{34}, spherical joints decouple the dynamic model of the system to translational and rotational dynamics, including aerodynamic forces and torques.

\subsection{Euler-Lagrange}

In the cable driven category, in \cite{31}, the Lagrange equations are derived for the quadrotor-pendulum model, considering both components as point masses with their positions parameterized. A point-mass payload is also considered in \cite{41}, \cite{115} and in \cite{119}. In \cite{106}, the control inputs are decoupled from the system dynamics providing redundancy. Considering a rigid payload instead of a point-pass, authors in \cite{44}, \cite{42}, \cite{43} and \cite{113} develop a coordinate free dynamic model using the E-L formulation. Specifically, in \cite{43} and \cite{44} the cables are modeled as a system composed of an arbitrary number of serially connected cables; the assumption of taut and massless cables is relaxed (that reduce stability in the case of aggressive maneuvers and aerodynamic effects in real-world conditions). Aerodynamic effects are considered in \cite{115}. In \cite{113}, the equations of motion relate the orientation of the links to the acceleration of the payload. In \cite{119}, two auxiliary planes (i.e., in-plane and out-of-plane motion) are proposed that allow for a clear separation of motion. The E-L formulation considers the masses of the payload and the follower during modeling; the leader accounts for defining the relative positions of the payload, and the follower the relative position with respect to the leader. This, however, does not directly impact dynamic behaviour of the overall system.

In the ground-air category, in \cite{50}, the E-L method is followed to describe the dynamic relationship between the UAV, UGV and the object, with the assumption that friction forces are negligible. In \cite{51}, the equations of motion are derived for each aerial and ground manipulator considering that the passive joints efficiently decouple the aerial manipulator's rotational dynamics and external wrenches (i.e., resultant force of the manipulator arm). In \cite{51} and \cite{52}, the complete dynamic model of both aerial and ground manipulators holding an object is derived based on the E-L formulation. In \cite{23}, a small deformation of the beam is considered when model the transportation system of two quadrotors with flexible payloads. In \cite{49}, a teleoperation system consisting of a haptic device (master model) and a team of aerial robots attached to an object (slave model), is modeled following with the Euler-Lagrange approach.  

The E-L formulation is mostly followed to model aerial manipulation systems of higher complexity, which consist of multi-DoF arms \cite{18}, \cite{15}, \cite{16}, \cite{17}, \cite{108}. In \cite{17}, the dynamics of the aerial manipulator and the object are combined using the force distribution solution \cite{107} at each grasping point. In \cite{15} and \cite{18} the twist of the payload is added in the payload dynamics. In addition, instead of assuming uniform mass distribution \cite{16}, in \cite{11}, the vector between the end-effector and the payloads' center of mass is estimated for transportation of a payload with an arbitrary mass. In \cite{19}, the kinematic model of the grasp is presented considering the end-effector forces in the model (i.e., balancing the object’s dynamics and the contact forces due to interacting with the environment, and internal forces). In \cite{12}, the dynamic relation of a multirotor combined with a multi-DOF robotic arm is discussed, with the assumption that the robotic arm is considered as an exogenous system (i.e., controlled independently). In \cite{10}, Lagrange equations are derived to model a quadrotor-manipulator (QM) system with a multi-DOF serial link arm. Dynamics decomposition is applied to decouple the QM system - this offers advantages (e.g., no inertial, Coriolis and gravity coupling). In \cite{20}, the dynamics of the overall combined system, 2 DoF arm aerial manipulators holding a bar, are presented with the assumption of a uniform distributed mass on the rigid object. 

\subsection{Other modeling methods}

The cable-driven category is also dominant when combined modeling methods are developed. In \cite{29} and \cite{30}, the dynamic equations of motion are derived for several helicopters (dynamics of the whole system) and a point mass load using the Kane method. In \cite{39} and \cite{40}, the Udwadia–Kalaba equations are adopted to calculate the constraint forces (i.e., a function of the accelerations of all involved bodies). N-E equations represent the different UAV and payload dynamics; aerodynamic drag and lifting conditions are considered during modeling. For cooperative manipulation of a box with catenary robots \cite{60}, planar modelling is followed considering cable assumptions (i.e., non-stretchable, massless) and constraints (i.e., cables that cannot have arbitrary angles, specific threshold for contact points). The N-E formulation describes the quadrotors' and payload's dynamics considering the forces and torques applied by the catenary robots to the box, and the E-L for the transportation system. In \cite{38}, authors calculate the available wrench set on the cables' tension in different cases (i.e., with or without external disturbance) concluding that the related to the minimum tension, acceleration of UAVs, maximum thrust of UAVs, external disturbance and the configuration of tethers (i.e., significance on the robustness of payload transportation). The dynamics of the quadrotor and the payload (i.e., point mass) follow N-E formulation while the equations of motion of the whole system are calculated using Udwadia–Kalaba equations for simulation purposes. In \cite{58}, regarding re-configurable quadrotors, authors model the interactions between payload and UAVs as influenced by cable forces using N-E and E-L methods. When it comes to modeling vibrations of a flexible payload of a cooperative aerial system, \cite{22} and \cite{54} discuss Euler-Bernoulli and E-L methods including only transverse vibration and planar modelling. 

\begin{table}[h]
\centering
\caption{Summarizing Table of Dynamic Modelling Approaches\label{tab:my-table2}}
\footnotesize
\newcolumntype{C}{>{\centering\arraybackslash}X}
\begin{tabularx}{0.85\textwidth}{CC}
\toprule
\textbf{Reference} & \textbf{Dynamic Modelling} \\
\midrule
\multicolumn{2}{c}{\textbf{Rigidly-Attached}} \\
\midrule
{\cite{45}}{\cite{46}}{\cite{63}}{\cite{47}}{\cite{48}}{\cite{56}}{\cite{34}} & Newton-Euler \\
\cite{49} & Euler-Lagrange \\
\midrule
\multicolumn{2}{c}{\textbf{Flexible object}} \\
\midrule
\cite{21} & Newton-Euler \\
\cite{23} & Euler-Lagrange \\
{\cite{22}}{\cite{54}} & Euler-Bernoulli, Euler-Lagrange \\
\midrule
\multicolumn{2}{c}{\textbf{Ground-Air}} \\
\midrule
{\cite{50}}{\cite{52}}{\cite{51}}{\cite{55}}{\cite{111}} & Euler-Lagrange \\
{\cite{110}} & Newton-Euler \\
\midrule
\multicolumn{2}{c}{\textbf{Multi DoF-Arms}} \\
\midrule
{\cite{11}}{\cite{10}}{\cite{19}}{\cite{12}}{\cite{17}}{\cite{16}}{\cite{20}}{\cite{15}}{\cite{18}} & Euler-Lagrange \\
{\cite{65}} & Newton-Euler \\
{\cite{66}}{\cite{112}} & Recursive Newton-Euler \\
\midrule
\multicolumn{2}{c}{\textbf{Cable-Driven}} \\
\midrule
{\cite{69}}{\cite{26}}{\cite{27}}{\cite{28}}{\cite{32}}{\cite{57}}{\cite{67}}{\cite{109}}{\cite{114}}{\cite{122}}{\cite{123}}{\cite{124}}{\cite{118}} & Newton-Euler \\
{\cite{29}}{\cite{30}} & Kane method \\
{\cite{59}}{\cite{31}}{\cite{106}}{\cite{44}}{\cite{43}}{\cite{42}}{\cite{41}}{\cite{108}}{\cite{113}}{\cite{115}}{\cite{119}}{\cite{117}} & Euler-Lagrange \\
{\cite{39}}{\cite{40}}{\cite{38}} & Newton-Euler, Udwadia-Kalaba \\
{\cite{60}}{\cite{58}} & Newton-Euler, Euler-Lagrange \\
\bottomrule
\end{tabularx}
\end{table}

\section{Control}

Control and controller design for cooperative aerial manipulation could be studied from different perspectives. When aerial vehicles interact with the environment (i.e., payload manipulation), it is essential that they need to be able to control at the same time the position at the contact, and the interaction force, preserving the stability of the entire system as well as the precision. Therefore, the main approach lies in considering the cooperative manipulation task as a united system to be controlled.

\subsection{Manipulation/Transportation}

In respect to the task of manipulation/transportation of an object by a team of multirotors various control strategies are employed and will be presented in this subsection. The task is completed without external disturbances or perception techniques.  

Control under Linear–quadratic regulator (LQR) seems to handle deformation and vibrations of the payload regarding transportation/manipulation with flexible objects. To be specific, due to the deformation of the payload that alters the relative position and orientation of the quadrotors \cite{23}, authors discuss LQR controller implementation along with integral terms based on the linearized model. In \cite{21} a continuous-time, infinite-horizon LQR controller is developed; each vehicle gives two scalar inputs (i.e., thrust force along z-axis, and a roll moment around x-axis). Moreover, in \cite{54}, collaborative control strategy comes to solve the under-actuation of the heterogeneous system (i.e., the flexible payload's dynamic matrix has one rank input). The ground manipulator controller generates a slow yet fully-actuated trajectory as the fast under-actuated aerial robot's control performs thrust aligning (i.e., orientation controller) and vibration suppression (i.e., LQR).

Geometric control approach is adopted for the cooperative transportation/manipulation between heterogenous robots \cite{51} and human-robot comanipulation \cite{111}. Specifically in \cite{53}, the aerial robot leverages a low-level geometric controller with an external wrench estimator and an admittance filter to manipulate the bar. The entire system is controlled by an augmented non linear controller \cite{51} with disturbance observer. In addition, in \cite{111} a geometric attitude controller tracks the desired orientation of the quadrotor while a human operator supplies the system with a force vector maintaining both cable and payload in specific attitudes. Although the collaborative task was successful, authors mention that a human-like controller could further enhance tracking performance. Two separate control units are exploited in \cite{50}. The UGV steers the object to the desired position thought PD control while the UAV adopts a cascaded control approach; inner loop for the attitude and outer loop for the inclination of the object. In \cite{55} leader-follower MPCs are employed for the transportation task of the heterogeneous robots.

The category regarding cooperative transportation/manipulation with multi-DoF arms contemplates force distribution and compensation of internal forces by exploiting force control and applying admittance/impedance filters for managing external forces. Except for the low level PD control for tracking each quadcopters' commands, authors in \cite{65} implement a decentralized adaptive controller that has no prior knowledge of the payload's mass yet succeeds to estimate the total mass by employing a force consensus algorithm. Authors mention that the controller ensures equal mass distribution that leads to equal thrust between agents; increasing flight time. In \cite{10} a hierarchical control framework was carried out in three layers; object behavior design, optimal cooperative force distribution and admittance force control. Regarding the force control of aerial manipulators, authors discuss a friction-cone constraint to maintain the contact with the grasp object. Moreover, model-based force control is presented in \cite{66}. In the design, the payload stiffness is considered and assumptions regarding the rigid grip are relaxed. In \cite{19} a control scheme consisting of two impedance filters (i.e., object level and internal) reduces contact (i.e, object/environment) and internal forces. In \cite{12} a null space based behavioral (NSB) approach prioritize tasks and achieves safe manipulation. Each multirotor is controlled by a robust controller (i.e., inner loop controller) with an extended high-gain and disturbance observer including arm dynamics. A reference velocity planner (i.e., outer loop controller) considers internal force and unilateral constraints in a NSB hierarchy. Path planning \cite{17}\cite{20} and collision-free path planning \cite{16}\cite{18} controllers are also demonstrated in this category. Works of \cite{17}\cite{16} and \cite{18} adopted augmented adaptive sliding mode controllers. In addition, control design including a high level augmented adaptive sliding mode controller and a low level robust controller with disturbance observer is presented in \cite{20}. Sliding mode controller is also applied in \cite{11} by incorporating two estimation algorithms (i.e., relative distance estimation and dynamic parameter estimation). 
 
In the previously published work of \cite{45} regarding rigidly-attached multirotors, authors discuss a two norm optimal control, partially decentralized. This implementation demonstrated successful results regarding grasping, stabilization and three-dimensional trajectories. In \cite{64} pre-flight estimation of physical parameters of an unknown payload is addressed. Moreover, authors identify the optimal arrangement of the attachment positions on the payload whereas the flight control itself is realized through a cascaded PID controller. For a more universal control system, a hierarchical controller and a modified control command allocation algorithm on each quadrotor is adopted \cite{56}; effective for a team of quadrotors with different orientations. In \cite{47} a dynamics-based control addresses joint limit constraints and thrust saturations as a constrained optimization problem. A decentralized autonomous smooth switching controller (ASSC) demonstrated robustness against robot failures \cite{63}.

Many previously published works develop geometric control for the cooperative transportation/manipulation task of cable-driven multirotors \cite{113}\cite{41}\cite{42}\cite{43}\cite{44} demonstrating agile maneuvers. Admittance filters \cite{32} also succeed to preserve stability of the transportation system. In addition, adaptive controllers seem to compensate for unknown parameters \cite{60} and adverse conditions \cite{67}. In \cite{60} an adaptive controller overcomes the challenges of the box's mass, inertia tensor, and contact points. By exploiting the subtask coordination of the NSB controller (i.e., allowing simultaneously load tracking and optimizing quadrotors' position), authors in \cite{108} employ a null-space-based adaptive control. NSB controller ensures trajectory tracking, whereas adaptive controllers (for UAV and manipulator) in the inner loop consider dynamic uncertainties, interactive forces, and unknown load. Simple schemes as demonstrated in \cite{26} and \cite{61} by exploiting PID in quasi-static motion and  cascaded PID at reasonable speeds respectively, showed inability to damp out oscillations. In \cite{123} a NMPC drives the object in a hierarchical way; the system explores secondary tasks such as inter-robot separation and obstacle avoidance in all 6 DoF. Authors address real-time computational requirements and less complexity by including only the payload pose in the state vector. Authors of \cite{124} propose a force coordination control scheme with disturbance separation and estimation, where a force-consensus term is introduced in order to average the load distribution between the quadrotors. The nonholonimc leader-follower system of \cite{118} leverages backstepping control on the leader and a switching controller on the follower. The backstepping method consists of two controllers (i.e., the kinematics controller and the dynamics controller) and two UKFs filters for forces estimation. In \cite{106} a nonlinear hierarchical controller incorporates two subsystems (i.e., inner, outer). In the outer loop authors considered the spatial swing angle between the quadrotor and the load. This strategy seem to increase controller's anti-swing performance. Focusing on motion planning, authors in \cite{69} present a distributed trajectory optimization control with parallel computation on a decomposition scheme. The method still relies in a centralized approach. Load-leading collaborative scheme \cite{109} employs a guidance control law that generates a common desired velocity for the quadrotors under a decentralized formation controller. Each quadrotor holds an internal feedback controller. In \cite{120} an estimation-based formation control tracks the desired trajectories under tension disturbance. In \cite{68} the combination of motion disagreements (i.e., no need for global positions) with an incremental nonlinear dynamic inversion controller (INDI), which particularly tracks acceleration signals, allows the accurate analysis of forces and accelerations of the system.

\subsection{In the presence of wind}

Wind gusts and external disturbances are inevitable when experiments are held in outdoor conditions and can degrade the performance and precision of the transportation/manipulation task. However, the majority of previously publish works in cooperative aerial manipulation and its individual categories, as described in IV, do not include experiments in the presence of wind (i.e., flexible payload, ground-air, multi-DoF arms categories). Nevertheless, experiments do exist in rigidly-attached and cable-driven categories. 

An admittance controller with force estimator is developed for the transportation task of rigidly-attached quadrotors in the presence of wind \cite{34}. In addition, a centralized finite state machine (FSM) coordinates the lifting and landing maneuver. When quadrotors reach the desired altitude, the admittance controller is engaged on every slave agent and allows an operator to control the master. 

Cable-driven multirotors illustrated agility in windy conditions. Certain experiments demonstrated small external disturbances \cite{115}\cite{39}, artificial force \cite{37}, wind-like disturbances \cite{67} (i.e., foam plates on drones to generate opposite drag forces), light breeze 7 km/h \cite{57} or even wind gusts up to 14.4-21.6 km/h \cite{40} and 10.8-25.2 km/h \cite{121}. In \cite{115} a sliding mode-adaptive PID showed robustness in the presence of external disturbances. The distributed adaptive controller of \cite{39} presented superior results to the nonlinear H$\infty$ under disturbances and without considering payload dynamics. In \cite{67} an inner-outer loop control scheme is adopted; a kinematic formation controller generates velocity reference while an adaptive dynamic compensator reduces the velocity-tracking error (i.e., produced from the system dynamics). In \cite{37} an uncertainty-aware H$\infty$ requires a minimum number of uncertain kinematic parameters and ensures stability though gain-tuning. Success on disturbance rejection is presented in passivity-based controllers \cite{114}\cite{40}. In the control design in \cite{114}, there is no requirement for payload measurements or cable connections at the center of mass (COM). In \cite{40}, a passivity-based decentralized formation control strategy with internal feedback control is introduced, assuming that the UAVs communicate with each other. This method has illustrated successful results with unknown payloads in unknown wind environments. Linear controller with a two-stage EKF-based estimator \cite{119} succeeds to transport and stabilize the object under wind disturbances. The state of the payload was estimated through the cable tension as obtained by the inertial sensor. Load-leading control also exhibit stable performance in the presence of light wind \cite{57}; PID feedback controllers with acceleration feed-forward are implemented for both the payload and quadcopter. Strong winds during the experiment in \cite{30} are compensated by an orientation controller; authors propose the use of force sensors in the ropes, enhancing robustness against variations of system parameters and disturbances. For the controller design in \cite{38}, a decentralized controller based on fixed-time extended state observer (ESO) is adopted, taking into account only the position and satisfying the desired figure-eight trajectory. To enhance the disturbance rejection ability, authors in \cite{121} presented a linear model-based force control (i.e., backstepping and thrust direction controllers) along with force and torque disturbance observers; the system succeeds to compensate cables' tension, strong winds and unmodeled dynamics. In the same experiment, formation planning in obstacle environments using virtual structure is also demonstrated. In \cite{62} model predictive control achieves smooth performance after demonstrating outdoors and at the same time allowing agile maneuvers with high precision; less than 5cm position accuracy. 

\subsection{Vision-based}

In the current literature, motion capture systems such as VICON and OptiTrack are usually employed to provide highly precise position and orientation information (i.e., state estimation). However, these sensors are used in indoor environments and therefore they could limit the application task. On the contrary, onboard visual sensors can provide the input for the state estimation, object detection \cite{15} or detect (not yet in the literature)/ track \cite{59} the manipulated object. 

Visual impedance control \cite{110} demonstrated stability for the human-robot cotransportation task. Instead of measuring the velocities of both human and robot, a visual observer estimates the relative velocity between the two. Authors mention that directly measuring human's state could be challenging for real life applications (e.g. need for GPS/ motion capture systems). In \cite{112} a leader-follower visual scheme performs transportation with two quadrotors equipped with dexterous robotic arms in gazebo environment. Authors mention that the controller was developed to acquire a more adaptive behaviour by not directly avoid compensating disturbances. The adaptive controller of \cite{15}, estimates the mass and inertial properties of an unknown payload by exploiting an on-line estimator. Moreover, the onboard RGB-D camera allows the system to detect obstacles real-time. In \cite{46} a non-linear control approach using inertial sensing is adopted for the transportation task of rigidly-attached multirotors. Pose estimation is obtained from visual-inertial odometry (VIO). This information is then sent to the controller in order to update the robots' states using a linear measurement model. Considering cable-driven robots, in \cite{117} and \cite{31}, LQR control strategies demonstrate successful transportation without communication, as the follower tracks the leader after employing vision techniques from on-board sensing (i.e., tags on the leader). Likewise, in \cite{59}, markers attached on the payload contribute in perception and state estimation under a distributed vision-based control approach. Each quadrotor processes the information from the onboard sensor (i.e., VIO information) and estimates its cable direction and velocity in a distributed way. MPC \cite{62} allows object detection and state estimation using on-board sensors.

\subsection{Teleoperation}

Teleoperation in cooperative aerial manipulation has potential towards safety as the human operator can directly control the manipulation task and therefore intervene if necessary. Bilateral teleoperation where skilled human operators generate the desired trajectory in a precise and safe way while being provided with force feedback to increase situational awareness is presented in rigidly-attached \cite{49} and ground-air \cite{52} categories, either by conducting simulations or through laboratory experiments. However, as mentioned in \cite{5}, imperfect communication, time-varying delay, information losses or possible destabilizing effects remain unexplored.

In \cite{49}, the teleoperation scheme incorporates a master-slave model (i.e., human-robots) where the human generates the desired trajectory through a haptic device. The trajectory is then sent to the object pose controller in order to compute the desired wrench. From there, a force allocation algorithm distributes the desired forces to each quadcopter CoMs and later passed to the attitude-and-thrust controllers (one for each quadcopter). The human receives the inertia of the whole system as a feedback and through a repulsive viscoelastic virtual force can have a realistic sense of the obstacles. Nevertheless, tested only in simulation. Lab experiment with haptic feedback employed in \cite{52}; a centralized task planner commands decision making of the Tele-MAGMAS system. Reconfigurable multirotors \cite{58} engage a dual space control approach with tension distribution. The method does not require to specify cable forces apriori; internal motion of the moving anchors is designed to keep the direction of the cables fixed w.r.t. to the payload (i.e., facilitating teleoperation).

\begin{table}[h]
\caption{Comparing state-of-the-art control approaches\label{tab:my-table3}}
\footnotesize
\newcolumntype{C}{>{\centering\arraybackslash}X}
\newcolumntype{R}{>{\centering\arraybackslash}m{0.6cm}}
\newcolumntype{G}{>{\centering\arraybackslash}m{1.1cm}} 
\newcolumntype{N}{>{\centering\arraybackslash}m{2cm}} 
\begin{tabularx}{\textwidth}{R|G|NCCCCCCCCC}
        \hline
        \toprule
        \textbf{Ref.} &
        \textbf{Category} &
        \textbf{Control-Technique} &
        \textbf{3D Trajectories} &
        \textbf{Velocity-Accel.} &
        \textbf{Payload Tracking Precision} &
        \textbf{Payload Orientation} &
        \textbf{Tested Against Uncertainties} &
        \textbf{Tested Against Wind} &
        \textbf{Tested Outdoors} &
        \textbf{Robust Against Disturbances} &
        \textbf{Implem. Complexity} \\ \hline
        {\cite{67}}  &  \multirow{9}{=}{\textit{Cable-Driven}} & Adaptive & Yes & 1.6 m/s\textsuperscript{2} & Low & 2DoF & Yes & Yes & No & No & Low \\ \cline{1-1} \cline{3-12} 
        {\cite{37}} & & H$\infty$ & No & QS & High & 3DoF & Yes & No & No & No & Med \\ \cline{1-1} \cline{3-12} 
        {\cite{57}} & & Load-leading (Optimal) & No & QS & High & 3DoF & Yes & Yes & Yes & No & High \\ \cline{1-1} \cline{3-12}
        {\cite{61}} & & Cascaded PID & No & 1 m/s & Low & No & Yes & No & Yes & No & Low \\ \cline{1-1} \cline{3-12}
        {\cite{119}} & & Linear controller with EKF-based estimator & Yes & 1.4m/s & Med & No & Yes & Yes & Yes & Yes & Med \\ \cline{1-1} \cline{3-12}
        {\cite{121}} & & Backstepp. force control and force DO & Yes & QS & Med & No & Yes & Yes & No & Yes & High \\ \cline{1-1} \cline{3-12}
        {\cite{38}} & & ESO & No & QS & Low & No & Yes & Yes & Yes & Yes & Med \\ \cline{1-1} \cline{3-12}
        {\cite{114}} & & Passivity-based & No & QS & Low & No & Yes & Yes & No & Yes & Low \\ \cline{1-1} \cline{3-12}
        {\cite{123}} & & MPC & No & <1m/s & High & 6DoF & No & No & No & No & High \\ \hline
        {\cite{23}} & \textit{Flexible Object} & LQR & No & 0.13m/s & Med & No & Yes & No & No & Yes & Low \\ \hline
        {\cite{18}} & \textit{Multi-DoF Arms} & Augmented adaptive sliding mode & Yes & <1m/s & Med & No & Yes & No & No & Yes & High \\ \hline
        {\cite{110}} & \textit{Human-UAV} & Visual Impedance & No & 0.3m/s & High & No & Yes & No & No & Yes & Med \\ \hline
        {\cite{34}} & \multirow{2}{=}{\\ \textit{Rigidly-Attached}} & Admittance with force estimation & No & <1m/s & Med & No & Yes & Yes & Yes & Yes & Med \\ \cline{1-1} \cline{3-12} 
        {\cite{56}} & & Hierarchical with command allocation & No & 0.1m/s & High & No & No & No & No & No & High \\ \bottomrule \hline        
    \end{tabularx}
\end{table}

\section{Conclusions and Discussion}

This survey summarizes published research in cooperative aerial manipulation focusing on last decade publications. Cooperative aerial manipulation is a challenging research and development area, and research findings and results are very promising. However, the overall cooperative aerial manipulation system is nonlinear with complex and coupled dynamics. Robust controller design is challenging; efficient navigation, accurate manipulation, and gripper (grasping) capabilities are also essential issues that need to be addressed, tackled and overcome.

Table \ref{tab:my-table4}, summarizes the previously published research that was discussed in this survey. Focusing on real-time implementation, one can see that most prototypes have been tested mainly in indoor/lab environments where wind disturbance (that may affect the system's performance dramatically) is not accounted for, thus simplifying considerably implementation. In most of the reviewed research, a motion capture system is used for state estimation. Although outdoor demonstrations do exist, mainly in cable-driven systems, there is still a gap until such systems can eventually execute successfully a real-life scenario. Besides, industrial/construction applications require heavier and more complicated load structures; in published work that uses rods or boxes, the heavier load is 4kg.

Several different platforms (i.e., quadrotors, hexarotors, octarotors, etc.) with different configurations (i.e, 2-DoF arms, rigid extensions, grippers etc.) are employed for different manipulation/transportation tasks. Multirotors appear to offer versatile solutions, having advantages such as high speed, stability and robustness against adverse environments. An additional advantage is their potential for increased flight time leveraging additive power sources, whereas their stability remains intact in the event of rotor failure, ensuring system integrity. A trend gaining traction in literature is the adoption of tilted propellers actuation, recognized for its suitability in physical interaction tasks (e.g. maintenance, inspection tasks), close-to-real-life scenario experiments \cite{52}\cite{101} and resilience against external disturbances without necessitating orientation adjustments.

Table \ref{tab:my-table2} showcases the dominant modeling approaches. Specifically, within the category of multi-DoF robotic arms, the E-L formulation accurately represents these intricate systems, as evidenced by laboratory experiments. This formulation also effectively captures the dynamic characteristics and behaviors of ground-air systems, as validated by practical experiments. Conversely, for rigidly-attached multirotors, the N-E formulation is favored, with its effectiveness demonstrated through experimentation.

Inspired by \cite{67}, Table \ref{tab:my-table3} presents the state-of-the art control approaches in cooperative aerial manipulation categories. The introduced methods depend on the most recent experiment in each category. A comparison is held between the most important features of the controller performance. Emphasis is given in payload tracking precision (indicated as low when the position error is more that 10cm, medium when it is between 5cm and 10cm and high when it is less than 5cm). Low payload tracking precision may be caused due to lack of available data of full pose information, or measurements from the payload that are not included in the controller design, or due to lack of available sensory data related to the payload. Uncertainties that need to be considered include: model uncertainties; environmental disturbances; hardware issues; sensor noise; payload variability, and, communications delays. Implementation complexity depends on the overall control architecture and followed methodology. Simulation studies and comparisons are not included in the table due to the focus in real-time implementable tasks. There exist advantages and limitations in each summarized method. For example, velocities and accelerations far from quasi static (QS) motion follow low payload tracking precision. However, they have been demonstrated in 3D trajectories. High payload tracking experiments seem to tackle high implementation complexity. Additionally, one can see that recent control methods include the ability to suppress uncertainties and disturbances (like wind) while keeping complexity low.

\subsection{Future Challenges}

A challenging and still open research topic is investigating combined data driven and model-based, as well as learning-based controller designs that consider trade offs between implementation simplicity and precision and also consider computational complexity \cite{5}. In addition, controller designs are needed that overcome challenges such as wind disturbances, normal speed and full pose manipulation. For implementation and testing of real-life scenarios, it is important to consider simplified assumptions, particularly when dealing with unknown values of a subset of system parameters. Regarding autonomy, onboard maps for navigating in unknown environments incur significant computational costs, yet it remains essential to facilitate adaptation capabilities when it comes to external disturbances. 

Prototypes found in the literature introduce quadrotors as the primary UAV platform for aerial manipulation, as quadrotors offer hardware simplicity. Nevertheless, when dealing with construction tasks that demand high autonomy, payload capacity, and increased flight time, their applicability is limited. Hyper-redundant dual arms are desirable but rarely used; they offer benefits like enhanced lifting capabilities and the ability to perform complex tasks within a larger workspace. But such redundancy has challenges due to added weights, potentially impacting autonomy. 

Although indoor and outdoor experiments do exist, most experiments still rely on external motion capture systems, prompting the need for future advances in perception and on-board processing. Safety is a notable constraint, frequently disregarded in existing literature. While effective obstacle avoidance in unknown environments has been showcased, the realm of cooperative aerial manipulation presents additional complexities, especially regarding potential human interaction. This underscores the importance of cautious consideration in future endeavors.

In conclusion, published literature reveals that despite notable progress in recent times, the development of a fully operational cooperative aerial manipulation system that is simultaneously safe, responsive, accurate, and capable of autonomous navigation solely through on-board visual sensors and local communication links (between neighbour robots) remains a challenge. System requirements delineate a highly promising yet formidable research domain within aerial robotics. A plethora of complex research challenges persists, necessitating resolution before such systems can achieve practical relevance and deployment in real-world scenarios, especially those demanding hard real-time capabilities.

\vspace{7pt} 

\textbf{Acknowledgments: }This article showcases the extended version of our previously published work "Real-Time Applicable Cooperative Aerial Manipulation: A Survey" (\href{https://ieeexplore.ieee.org/document/10155960}{CrossRef}) which was presented at the 2023 International Conference on Unmanned Aircraft Systems (ICUAS).

\textbf{Author Contributions: }All authors contributed to the study conception and design. Material preparation, data collection and analysis were performed by S.C.B., C.S.T. and K.P.V. The first draft of the manuscript was written by S.C.B., and all authors commented on previous versions of the manuscript. All authors have read and agreed to the published version of the manuscript.

\textbf{Funding: }The work in this paper has been supported by Special Account for Research Funding (E.L.K.E.) of National Technical University of Athens (N.T.U.A.).

\textbf{Conflicts of interest: }No known conflicts of interest.

\begin{table}[]
\caption{Summarizing Table of Cooperative Aerial Manipulation's different categories}
\label{tab:my-table4}
\huge
\resizebox{\textwidth}{!}{%
\begin{tabular}{|cccccccccccccc|}
\hline
\multicolumn{1}{|c|}{\textit{Ref.}} &
  \multicolumn{1}{c|}{\textit{\begin{tabular}[c]{@{}c@{}}Fixed/\\ Tilted\end{tabular}}} &
  \multicolumn{2}{c|}{\textit{Arms}} &
  \multicolumn{1}{c|}{\textit{Gripper Type}} &
  \multicolumn{1}{c|}{\textit{Platform}} &
  \multicolumn{1}{c|}{\textit{Implementation}} &
  \multicolumn{1}{c|}{\textit{UAV Weight}} &
  \multicolumn{1}{c|}{\textit{Payload Weight}} &
  \multicolumn{1}{c|}{\textit{Payload}} &
  \multicolumn{1}{c|}{\textit{Controller Design}} &
  \multicolumn{1}{c|}{\textit{\begin{tabular}[c]{@{}c@{}}Motion Planning \\ Technique\end{tabular}}} &
  \multicolumn{1}{c|}{\textit{Fully Autonomous}} &
  \textit{\begin{tabular}[c]{@{}c@{}}Experiment \\ Task\end{tabular}} \\ \hline
\multicolumn{14}{|c|}{\textbf{RIGIDLY ATTACHED}} \\ \hline
\multicolumn{1}{|c|}{{\cite{34}}} &
  \multicolumn{1}{c|}{} &
  \multicolumn{2}{c|}{\begin{tabular}[c]{@{}c@{}}Gripper with\\ spherical joint\end{tabular}} &
  \multicolumn{1}{c|}{} &
  \multicolumn{1}{c|}{Hexacopter} &
  \multicolumn{1}{c|}{Outdoor experiment} &
  \multicolumn{1}{c|}{1kg} &
  \multicolumn{1}{c|}{2.46kg} &
  \multicolumn{1}{c|}{Wooden structure} &
  \multicolumn{1}{c|}{\begin{tabular}[c]{@{}c@{}}Admittance controller \\ with force estimator\end{tabular}} &
  \multicolumn{1}{c|}{} &
  \multicolumn{1}{c|}{\begin{tabular}[c]{@{}c@{}}Onboard   \\ sensors\end{tabular}} &
   \\ \cline{1-1} \cline{3-4} \cline{6-11} \cline{13-13}
\multicolumn{1}{|c|}{{\cite{46}}} &
  \multicolumn{1}{c|}{} &
  \multicolumn{2}{c|}{} &
  \multicolumn{1}{c|}{\multirow{-2}{*}{Magnetic}} &
  \multicolumn{1}{c|}{} &
  \multicolumn{1}{c|}{} &
  \multicolumn{2}{c|}{610gr} &
  \multicolumn{1}{c|}{Carbon fiber rod} &
  \multicolumn{1}{c|}{Non-Linear Controller} &
  \multicolumn{1}{c|}{\multirow{-2}{*}{-}} &
  \multicolumn{1}{c|}{\begin{tabular}[c]{@{}c@{}}Onboard  \\ camera \\ and IMU\end{tabular}} &
  \multirow{-2}{*}{Transportation} \\ \cline{1-1} \cline{5-5} \cline{8-14} 
\multicolumn{1}{|c|}{{\cite{45}}} &
  \multicolumn{1}{c|}{} &
  \multicolumn{2}{c|}{\multirow{-2}{*}{Gripper}} &
  \multicolumn{1}{c|}{Penetrative} &
  \multicolumn{1}{c|}{\multirow{-2}{*}{Quadcopter}} &
  \multicolumn{1}{c|}{} &
  \multicolumn{1}{c|}{500gr} &
  \multicolumn{1}{c|}{320gr} &
  \multicolumn{1}{c|}{Wooden bar} &
  \multicolumn{1}{c|}{\begin{tabular}[c]{@{}c@{}}Two norm optimal control, \\ partially decentralized\end{tabular}} &
  \multicolumn{2}{c|}{} &
  \begin{tabular}[c]{@{}c@{}}Pick/\\ Transportation\end{tabular} \\ \cline{1-1} \cline{3-6} \cline{8-11} \cline{14-14} 
\multicolumn{1}{|c|}{{\cite{63}}} &
  \multicolumn{1}{c|}{} &
  \multicolumn{2}{c|}{} &
  \multicolumn{1}{c|}{\begin{tabular}[c]{@{}c@{}}Fixed position \\ on the payload\end{tabular}} &
  \multicolumn{1}{c|}{} &
  \multicolumn{1}{c|}{} &
  \multicolumn{2}{c|}{2.7kg} &
  \multicolumn{1}{c|}{\begin{tabular}[c]{@{}c@{}}Long rectangular \\ object\end{tabular}} &
  \multicolumn{1}{c|}{\begin{tabular}[c]{@{}c@{}}Decentralized controller \\ based ASSC\end{tabular}} &
  \multicolumn{2}{c|}{} &
   \\ \cline{1-1} \cline{5-5} \cline{8-11}
\multicolumn{1}{|c|}{{\cite{64}}} &
  \multicolumn{1}{c|}{} &
  \multicolumn{2}{c|}{\multirow{-2}{*}{-}} &
  \multicolumn{1}{c|}{\begin{tabular}[c]{@{}c@{}}Variable attachment \\ mechanism \\ on the payload\end{tabular}} &
  \multicolumn{1}{c|}{\multirow{-2}{*}{Mono-rotor}} &
  \multicolumn{1}{c|}{\multirow{-4}{*}{Lab experiment}} &
  \multicolumn{2}{c|}{3.2kg} &
  \multicolumn{1}{c|}{\begin{tabular}[c]{@{}c@{}}Long  rectangular\\ object\end{tabular}} &
  \multicolumn{1}{c|}{Cascaded PID} &
  \multicolumn{2}{c|}{} &
  \multirow{-2}{*}{Transportation} \\ \cline{1-1} \cline{3-11} \cline{14-14} 
\multicolumn{1}{|c|}{{\cite{49}}} &
  \multicolumn{1}{c|}{} &
  \multicolumn{2}{c|}{Passive tool} &
  \multicolumn{1}{c|}{Adhesive tooltip} &
  \multicolumn{1}{c|}{} &
  \multicolumn{1}{c|}{\begin{tabular}[c]{@{}c@{}}Simulation with \\ Hardware in the loop\end{tabular}} &
  \multicolumn{1}{c|}{0.7kg} &
  \multicolumn{1}{c|}{2.4kg} &
  \multicolumn{1}{c|}{Barrel} &
  \multicolumn{1}{c|}{Haptic feedback control} &
  \multicolumn{2}{c|}{} &
  Manipulation \\ \cline{1-1} \cline{3-5} \cline{7-11} \cline{14-14} 
\multicolumn{1}{|c|}{{\cite{47}}} &
  \multicolumn{1}{c|}{} &
  \multicolumn{2}{c|}{\begin{tabular}[c]{@{}c@{}}Spherically connected \\ frame with \\ attached rigid tool\end{tabular}} &
  \multicolumn{1}{c|}{-} &
  \multicolumn{1}{c|}{} &
  \multicolumn{1}{c|}{} &
  \multicolumn{1}{c|}{\begin{tabular}[c]{@{}c@{}}2.31kg   \\ (3 quads)\end{tabular}} &
  \multicolumn{1}{c|}{1.2kg} &
  \multicolumn{1}{c|}{Box} &
  \multicolumn{1}{c|}{\begin{tabular}[c]{@{}c@{}}Dynamics-Based \\ Control\end{tabular}} &
  \multicolumn{2}{c|}{} &
  \begin{tabular}[c]{@{}c@{}}Point-Contact/   \\ Push\end{tabular} \\ \cline{1-1} \cline{3-5} \cline{8-11} \cline{14-14} 
\multicolumn{1}{|c|}{{\cite{56}}} &
  \multicolumn{1}{c|}{\multirow{-8}{*}{Fixed}} &
  \multicolumn{2}{c|}{-} &
  \multicolumn{1}{c|}{Magnets} &
  \multicolumn{1}{c|}{\multirow{-3}{*}{Quadcopter}} &
  \multicolumn{1}{c|}{\multirow{-2}{*}{Lab experiment}} &
  \multicolumn{1}{c|}{1.121kg} &
  \multicolumn{1}{c|}{0.453kg} &
  \multicolumn{1}{c|}{Wooden board} &
  \multicolumn{1}{c|}{\begin{tabular}[c]{@{}c@{}}Hierarchical control \\ with command allocation\end{tabular}} &
  \multicolumn{2}{c|}{\multirow{-6}{*}{-}} &
  Transportation \\ \hline
\multicolumn{14}{|c|}{\textbf{FLEXIBLE OBJECTS}} \\ \hline
\multicolumn{1}{|c|}{{\cite{21}}} &
  \multicolumn{1}{c|}{} &
  \multicolumn{2}{c|}{} &
  \multicolumn{1}{c|}{\begin{tabular}[c]{@{}c@{}}Attached   \\ to the payload\end{tabular}} &
  \multicolumn{1}{c|}{Quadcopter} &
  \multicolumn{1}{c|}{} &
  \multicolumn{1}{c|}{0.47kg} &
  \multicolumn{1}{c|}{0.54   kg} &
  \multicolumn{1}{c|}{Aluminum ring} &
  \multicolumn{1}{c|}{LQR} &
  \multicolumn{2}{c|}{} &
   \\ \cline{1-1} \cline{5-6} \cline{8-11}
\multicolumn{1}{|c|}{{\cite{23}}} &
  \multicolumn{1}{c|}{} &
  \multicolumn{2}{c|}{} &
  \multicolumn{1}{c|}{\begin{tabular}[c]{@{}c@{}}Attached   \\ to the payload \\ via magnets\end{tabular}} &
  \multicolumn{1}{c|}{Octacopter} &
  \multicolumn{1}{c|}{\multirow{-2}{*}{Lab experiment}} &
  \multicolumn{2}{c|}{1.121kg} &
  \multicolumn{1}{c|}{Beam} &
  \multicolumn{1}{c|}{LQR} &
  \multicolumn{2}{c|}{} &
  \multirow{-2}{*}{Transportation} \\ \cline{1-1} \cline{5-11} \cline{14-14} 
\multicolumn{1}{|c|}{{\cite{54}}} &
  \multicolumn{1}{c|}{\multirow{-3}{*}{Fixed}} &
  \multicolumn{2}{c|}{} &
  \multicolumn{1}{c|}{\begin{tabular}[c]{@{}c@{}}Attached  link to \\ payload through \\ passive rotational joint\end{tabular}} &
  \multicolumn{1}{c|}{Quadcopter} &
  \multicolumn{1}{c|}{Simulation} &
  \multicolumn{1}{c|}{Not reported} &
  \multicolumn{1}{c|}{0.61kg} &
  \multicolumn{1}{c|}{Wooden bar} &
  \multicolumn{1}{c|}{\begin{tabular}[c]{@{}c@{}}Collaborative control \\ with vibration suspension\end{tabular}} &
  \multicolumn{2}{c|}{} &
   \\ \cline{1-2} \cline{5-11}
\multicolumn{1}{|c|}{{\cite{22}}} &
  \multicolumn{1}{c|}{Tilted} &
  \multicolumn{2}{c|}{\multirow{-4}{*}{-}} &
  \multicolumn{1}{c|}{\begin{tabular}[c]{@{}c@{}}Attached   \\ to the payload\end{tabular}} &
  \multicolumn{1}{c|}{Two-rotor} &
  \multicolumn{1}{c|}{Lab experiment} &
  \multicolumn{2}{c|}{Not reported} &
  \multicolumn{1}{c|}{Aluminum bar} &
  \multicolumn{1}{c|}{LQR} &
  \multicolumn{2}{c|}{\multirow{-4}{*}{-}} &
  \multirow{-2}{*}{Manipulation} \\ \hline
\multicolumn{14}{|c|}{\textbf{GROUND-AIR}} \\ \hline
\multicolumn{1}{|c|}{{\cite{50}}} &
  \multicolumn{1}{c|}{} &
  \multicolumn{2}{c|}{} &
  \multicolumn{1}{c|}{\begin{tabular}[c]{@{}c@{}}Attached   \\ to the payload\end{tabular}} &
  \multicolumn{1}{c|}{Birotor} &
  \multicolumn{1}{c|}{} &
  \multicolumn{1}{c|}{100gr} &
  \multicolumn{1}{c|}{30gr} &
  \multicolumn{1}{c|}{Carbon rod} &
  \multicolumn{1}{c|}{PD} &
  \multicolumn{2}{c|}{} &
   \\ \cline{1-1} \cline{5-6} \cline{8-11}
\multicolumn{1}{|c|}{{\cite{51}}} &
  \multicolumn{1}{c|}{} &
  \multicolumn{2}{c|}{\multirow{-2}{*}{-}} &
  \multicolumn{1}{c|}{\begin{tabular}[c]{@{}c@{}}Attached link to payload\\  through passive \\ rotational joint\end{tabular}} &
  \multicolumn{1}{c|}{Quadcopter} &
  \multicolumn{1}{c|}{} &
  \multicolumn{1}{c|}{Not reported} &
  \multicolumn{1}{c|}{0.61kg} &
  \multicolumn{1}{c|}{Wooden bar} &
  \multicolumn{1}{c|}{\begin{tabular}[c]{@{}c@{}}Augmented nonlinear \\ with disturbance observer\end{tabular}} &
  \multicolumn{2}{c|}{} &
  \multirow{-2}{*}{Manipulation} \\ \cline{1-1} \cline{3-6} \cline{8-11} \cline{14-14} 
\multicolumn{1}{|c|}{{\cite{55}}} &
  \multicolumn{1}{c|}{\multirow{-3}{*}{Fixed}} &
  \multicolumn{2}{c|}{2-DOF arm (x1)} &
  \multicolumn{1}{c|}{Gripper} &
  \multicolumn{1}{c|}{} &
  \multicolumn{1}{c|}{\multirow{-3}{*}{Lab experiment}} &
  \multicolumn{1}{c|}{Not reported} &
  \multicolumn{1}{c|}{Not reported} &
  \multicolumn{1}{c|}{Plastic bar} &
  \multicolumn{1}{c|}{Model Predictive Control} &
  \multicolumn{2}{c|}{} &
  \begin{tabular}[c]{@{}c@{}}Manipulation/\\ Transportation\end{tabular} \\ \cline{1-5} \cline{7-11} \cline{14-14} 
\multicolumn{1}{|c|}{{\cite{52}}} &
  \multicolumn{1}{c|}{Tilted} &
  \multicolumn{2}{c|}{\begin{tabular}[c]{@{}c@{}}1-DOF arm (x1) \\ with passive joint\end{tabular}} &
  \multicolumn{1}{c|}{Gripper (x2)} &
  \multicolumn{1}{c|}{\multirow{-2}{*}{Hexacopter}} &
  \multicolumn{1}{c|}{Indoor experiment} &
  \multicolumn{1}{c|}{2.48kg} &
  \multicolumn{1}{c|}{1.42kg} &
  \multicolumn{1}{c|}{Rod} &
  \multicolumn{1}{c|}{\begin{tabular}[c]{@{}c@{}}Centralized controller \\ and haptic feedback\end{tabular}} &
  \multicolumn{2}{c|}{\multirow{-4}{*}{-}} &
  \begin{tabular}[c]{@{}c@{}}Pick/\\ Manipulation\end{tabular} \\ \hline
\multicolumn{1}{|c|}{{\cite{111}}} &
  \multicolumn{1}{c|}{} &
  \multicolumn{2}{c|}{} &
  \multicolumn{1}{c|}{} &
  \multicolumn{1}{c|}{Quadcopter} &
  \multicolumn{1}{c|}{Outdoor experiment} &
  \multicolumn{1}{c|}{1.236kg} &
  \multicolumn{1}{c|}{0.26kg} &
  \multicolumn{1}{c|}{Rod} &
  \multicolumn{1}{c|}{Collaborative control} &
  \multicolumn{1}{c|}{} &
  \multicolumn{1}{c|}{\begin{tabular}[c]{@{}c@{}}GPS, IMU, \\ custom sensors\\ system\end{tabular}} &
  \begin{tabular}[c]{@{}c@{}}Pick/Manipulation/\\ Transportation\end{tabular} \\ \cline{1-1} \cline{6-11} \cline{13-14} 
\multicolumn{1}{|c|}{{\cite{110}}} &
  \multicolumn{1}{c|}{\multirow{-2}{*}{Fixed}} &
  \multicolumn{2}{c|}{\multirow{-2}{*}{Cable}} &
  \multicolumn{1}{c|}{\multirow{-2}{*}{-}} &
  \multicolumn{1}{c|}{Hexacopter} &
  \multicolumn{1}{c|}{Lab experiment} &
  \multicolumn{1}{c|}{4.05kg} &
  \multicolumn{1}{c|}{1.12kg} &
  \multicolumn{1}{c|}{Bar} &
  \multicolumn{1}{c|}{Visual Impedance control} &
  \multicolumn{1}{c|}{\multirow{-2}{*}{-}} &
  \multicolumn{1}{c|}{\begin{tabular}[c]{@{}c@{}}Onboard camera \\ and IMU\end{tabular}} &
  \begin{tabular}[c]{@{}c@{}}Manipulation/\\ Transportation\end{tabular} \\ \hline
\multicolumn{14}{|c|}{\textbf{MULTI-DOF ARMS}} \\ \hline
\multicolumn{1}{|c|}{{\cite{12}}} &
  \multicolumn{1}{c|}{} &
  \multicolumn{2}{c|}{3-DOF arm (x1)} &
  \multicolumn{1}{c|}{\begin{tabular}[c]{@{}c@{}}Connected to the object \\ using ball joints\end{tabular}} &
  \multicolumn{1}{c|}{} &
  \multicolumn{1}{c|}{} &
  \multicolumn{1}{c|}{3kg} &
  \multicolumn{1}{c|}{500gr} &
  \multicolumn{1}{c|}{} &
  \multicolumn{1}{c|}{\begin{tabular}[c]{@{}c@{}}Robust controller \\ with disturbance observer\end{tabular}} &
  \multicolumn{2}{c|}{-} &
   \\ \cline{1-1} \cline{3-5} \cline{8-9} \cline{11-13}
\multicolumn{1}{|c|}{{\cite{17}}} &
  \multicolumn{1}{c|}{} &
  \multicolumn{2}{c|}{2-DOF arm (x1)} &
  \multicolumn{1}{c|}{Hook} &
  \multicolumn{1}{c|}{} &
  \multicolumn{1}{c|}{} &
  \multicolumn{1}{c|}{} &
  \multicolumn{1}{c|}{150gr} &
  \multicolumn{1}{c|}{} &
  \multicolumn{1}{c|}{} &
  \multicolumn{1}{c|}{RRT*} &
  \multicolumn{1}{c|}{-} &
   \\ \cline{1-1} \cline{3-5} \cline{9-9} \cline{12-13}
\multicolumn{1}{|c|}{{\cite{16}}} &
  \multicolumn{1}{c|}{} &
  \multicolumn{2}{c|}{2-DOF arm (x1)} &
  \multicolumn{1}{c|}{Gripper} &
  \multicolumn{1}{c|}{} &
  \multicolumn{1}{c|}{} &
  \multicolumn{1}{c|}{} &
  \multicolumn{1}{c|}{280gr} &
  \multicolumn{1}{c|}{} &
  \multicolumn{1}{c|}{\multirow{-2}{*}{\begin{tabular}[c]{@{}c@{}}Augmented adaptive sliding \\ mode controller\end{tabular}}} &
  \multicolumn{1}{c|}{\begin{tabular}[c]{@{}c@{}}RRT* with Bezier \\ Curves and DMPs\end{tabular}} &
  \multicolumn{1}{c|}{-} &
   \\ \cline{1-1} \cline{3-5} \cline{9-9} \cline{11-13}
\multicolumn{1}{|c|}{{\cite{20}}} &
  \multicolumn{1}{c|}{} &
  \multicolumn{2}{c|}{2-DOF arm (x1)} &
  \multicolumn{1}{c|}{Rigidly attached} &
  \multicolumn{1}{c|}{} &
  \multicolumn{1}{c|}{} &
  \multicolumn{1}{c|}{} &
  \multicolumn{1}{c|}{280gr} &
  \multicolumn{1}{c|}{\multirow{-4}{*}{Rod}} &
  \multicolumn{1}{c|}{\begin{tabular}[c]{@{}c@{}}High level augmented \\ adaptive sliding mode controller \\ and low level robust controller \\ with DOB\end{tabular}} &
  \multicolumn{1}{c|}{\begin{tabular}[c]{@{}c@{}}Learning based \\ PDMPs\end{tabular}} &
  \multicolumn{1}{c|}{-} &
   \\ \cline{1-1} \cline{3-5} \cline{9-13}
\multicolumn{1}{|c|}{{\cite{15}}} &
  \multicolumn{1}{c|}{} &
  \multicolumn{2}{c|}{2-DOF arm (x1)} &
  \multicolumn{1}{c|}{Gripper} &
  \multicolumn{1}{c|}{} &
  \multicolumn{1}{c|}{} &
  \multicolumn{1}{c|}{\multirow{-4}{*}{Not reported}} &
  \multicolumn{1}{c|}{280gr} &
  \multicolumn{1}{c|}{Wooden rod} &
  \multicolumn{1}{c|}{Adaptive controller} &
  \multicolumn{1}{c|}{DMPs} &
  \multicolumn{1}{c|}{\begin{tabular}[c]{@{}c@{}}RGB-D camera \\ for object detection\end{tabular}} &
   \\ \cline{1-1} \cline{3-5} \cline{8-13}
\multicolumn{1}{|c|}{{\cite{18}}} &
  \multicolumn{1}{c|}{} &
  \multicolumn{2}{c|}{2-DOF arm (x1)} &
  \multicolumn{1}{c|}{} &
  \multicolumn{1}{c|}{\multirow{-6}{*}{Hexacopter}} &
  \multicolumn{1}{c|}{\multirow{-6}{*}{Lab experiment}} &
  \multicolumn{1}{c|}{2.5kg} &
  \multicolumn{1}{c|}{280gr} &
  \multicolumn{1}{c|}{Rod} &
  \multicolumn{1}{c|}{\begin{tabular}[c]{@{}c@{}}Augmented adaptive \\ sliding mode controller\end{tabular}} &
  \multicolumn{1}{c|}{\begin{tabular}[c]{@{}c@{}}Velocity and \\ curvature\\ constraints\end{tabular}} &
  \multicolumn{1}{c|}{-} &
   \\ \cline{1-1} \cline{3-4} \cline{6-13}
\multicolumn{1}{|c|}{{\cite{11}}} &
  \multicolumn{1}{c|}{} &
  \multicolumn{2}{c|}{2-DOF arm (x1)} &
  \multicolumn{1}{c|}{} &
  \multicolumn{1}{c|}{} &
  \multicolumn{1}{c|}{} &
  \multicolumn{1}{c|}{Not   reported} &
  \multicolumn{1}{c|}{50gr} &
  \multicolumn{1}{c|}{\begin{tabular}[c]{@{}c@{}}Rod/triangular   \\ object\end{tabular}} &
  \multicolumn{1}{c|}{\begin{tabular}[c]{@{}c@{}}Sliding mode controller \\ with SOSM observer\end{tabular}} &
  \multicolumn{2}{c|}{} &
  \multirow{-7}{*}{Manipulation} \\ \cline{1-1} \cline{3-4} \cline{8-11} \cline{14-14} 
\multicolumn{1}{|c|}{{\cite{10}}} &
  \multicolumn{1}{c|}{} &
  \multicolumn{2}{c|}{2-DOF arm (x1)} &
  \multicolumn{1}{c|}{\multirow{-3}{*}{Rigidly attached}} &
  \multicolumn{1}{c|}{\multirow{-2}{*}{Quadcopter}} &
  \multicolumn{1}{c|}{} &
  \multicolumn{1}{c|}{2kg} &
  \multicolumn{1}{c|}{0.4kg} &
  \multicolumn{1}{c|}{Box/cylinder} &
  \multicolumn{1}{c|}{\begin{tabular}[c]{@{}c@{}}Hierarchical with \\ admittance force control\end{tabular}} &
  \multicolumn{2}{c|}{\multirow{-2}{*}{-}} &
  \begin{tabular}[c]{@{}c@{}}Push/\\ Transportation\end{tabular} \\ \cline{1-1} \cline{3-6} \cline{8-14} 
\multicolumn{1}{|c|}{{\cite{19}}} &
  \multicolumn{1}{c|}{} &
  \multicolumn{2}{c|}{6-DOF arm (x1)} &
  \multicolumn{1}{c|}{Gripper} &
  \multicolumn{1}{c|}{Octacopter} &
  \multicolumn{1}{c|}{\multirow{-3}{*}{Simulation}} &
  \multicolumn{1}{c|}{2.7kg} &
  \multicolumn{1}{c|}{1kg} &
  \multicolumn{1}{c|}{Rod} &
  \multicolumn{1}{c|}{Impedance control} &
  \multicolumn{1}{c|}{-} &
  \multicolumn{1}{c|}{-} &
  Transportation \\ \cline{1-1} \cline{3-14} 
\multicolumn{1}{|c|}{{\cite{112}}} &
  \multicolumn{1}{c|}{} &
  \multicolumn{2}{c|}{2-DOF arm (x1)} &
  \multicolumn{1}{c|}{Virtual extension} &
  \multicolumn{1}{c|}{} &
  \multicolumn{1}{c|}{Gazebo simulation} &
  \multicolumn{1}{c|}{Not reported} &
  \multicolumn{1}{c|}{Not reported} &
  \multicolumn{1}{c|}{Not reported} &
  \multicolumn{1}{c|}{Compliant controller} &
  \multicolumn{1}{c|}{-} &
  \multicolumn{1}{c|}{\begin{tabular}[c]{@{}c@{}}RGB-camera \\ (visual feedback \\ using markers)\end{tabular}} &
  \begin{tabular}[c]{@{}c@{}}Manipulation/\\ Transportation\end{tabular} \\ \cline{1-1} \cline{3-5} \cline{7-14} 
\multicolumn{1}{|c|}{{\cite{65}}} &
  \multicolumn{1}{c|}{} &
  \multicolumn{2}{c|}{Rigid extension} &
  \multicolumn{1}{c|}{Rigidly attached} &
  \multicolumn{1}{c|}{} &
  \multicolumn{1}{c|}{Simulation} &
  \multicolumn{1}{c|}{0.75kg} &
  \multicolumn{1}{c|}{1.5kg} &
  \multicolumn{1}{c|}{Spherical object} &
  \multicolumn{1}{c|}{\begin{tabular}[c]{@{}c@{}}Decentralized adaptive \\ force control\end{tabular}} &
  \multicolumn{2}{c|}{} &
   \\ \cline{1-1} \cline{3-5} \cline{7-11}
\multicolumn{1}{|c|}{{\cite{66}}} &
  \multicolumn{1}{c|}{\multirow{-12}{*}{Fixed}} &
  \multicolumn{2}{c|}{2-DOF arm (x1)} &
  \multicolumn{1}{c|}{Gripper} &
  \multicolumn{1}{c|}{\multirow{-3}{*}{Quadcopter}} &
  \multicolumn{1}{c|}{Gazebo simulation} &
  \multicolumn{1}{c|}{1.2kg} &
  \multicolumn{1}{c|}{0.8kg} &
  \multicolumn{1}{c|}{Beam} &
  \multicolumn{1}{c|}{\begin{tabular}[c]{@{}c@{}}Leader/Follower   \\ model-based force control\end{tabular}} &
  \multicolumn{2}{c|}{\multirow{-2}{*}{-}} &
  \multirow{-2}{*}{Transportation} \\ \hline
\multicolumn{14}{|c|}{\textbf{CABLE-DRIVEN}} \\ \hline
\multicolumn{1}{|c|}{{\cite{26}}} &
  \multicolumn{1}{c|}{} &
  \multicolumn{3}{c|}{} &
  \multicolumn{1}{c|}{} &
  \multicolumn{1}{c|}{} &
  \multicolumn{1}{c|}{Not reported} &
  \multicolumn{1}{c|}{0.25kg} &
  \multicolumn{1}{c|}{Triangular object} &
  \multicolumn{1}{c|}{PID} &
  \multicolumn{2}{c|}{} &
   \\ \cline{1-1} \cline{8-11}
\multicolumn{1}{|c|}{{\cite{37}}} &
  \multicolumn{1}{c|}{} &
  \multicolumn{3}{c|}{} &
  \multicolumn{1}{c|}{\multirow{-2}{*}{Quadcopter}} &
  \multicolumn{1}{c|}{\multirow{-2}{*}{Lab experiment}} &
  \multicolumn{1}{c|}{1.03kg} &
  \multicolumn{1}{c|}{0.338kg} &
  \multicolumn{1}{c|}{\begin{tabular}[c]{@{}c@{}}Platform of carbon \\ fiber bars\end{tabular}} &
  \multicolumn{1}{c|}{Robust controller} &
  \multicolumn{2}{c|}{\multirow{-2}{*}{-}} &
  \multirow{-2}{*}{\begin{tabular}[c]{@{}c@{}}Lift/Manipulation/\\ Transportation\end{tabular}} \\ \cline{1-1} \cline{6-14} 
\multicolumn{1}{|c|}{{\cite{30}}} &
  \multicolumn{1}{c|}{} &
  \multicolumn{3}{c|}{\multirow{-3}{*}{-}} &
  \multicolumn{1}{c|}{Helicopter} &
  \multicolumn{1}{c|}{Outdoor experiment} &
  \multicolumn{1}{c|}{12.5kg} &
  \multicolumn{1}{c|}{4kg} &
  \multicolumn{1}{c|}{Camera} &
  \multicolumn{1}{c|}{Orientation controller} &
  \multicolumn{1}{c|}{-} &
  \multicolumn{1}{c|}{\begin{tabular}[c]{@{}c@{}}IMU, GPS, \\ Compass\end{tabular}} &
  Lift/Transportation \\ \cline{1-1} \cline{3-14} 
\multicolumn{1}{|c|}{{\cite{31}}} &
  \multicolumn{1}{c|}{} &
  \multicolumn{2}{c|}{-} &
  \multicolumn{1}{c|}{Magnetic grippers} &
  \multicolumn{1}{c|}{} &
  \multicolumn{1}{c|}{Lab experiment} &
  \multicolumn{1}{c|}{800gr} &
  \multicolumn{1}{c|}{263gr} &
  \multicolumn{1}{c|}{Aluminum rod} &
  \multicolumn{1}{c|}{LQR} &
  \multicolumn{1}{c|}{-} &
  \multicolumn{1}{c|}{\begin{tabular}[c]{@{}c@{}}Monocular camera, \\ IMU\end{tabular}} &
  Pick/Transportation \\ \cline{1-1} \cline{3-5} \cline{7-14} 
\multicolumn{1}{|c|}{{\cite{39}}} &
  \multicolumn{1}{c|}{} &
  \multicolumn{3}{c|}{} &
  \multicolumn{1}{c|}{} &
  \multicolumn{1}{c|}{Simulation} &
  \multicolumn{1}{c|}{1kg} &
  \multicolumn{1}{c|}{2.65kg} &
  \multicolumn{1}{c|}{Rectangular object} &
  \multicolumn{1}{c|}{Distributed adaptive control} &
  \multicolumn{1}{c|}{\begin{tabular}[c]{@{}c@{}}RRT* with \\ B-Spline curve\end{tabular}} &
  \multicolumn{1}{c|}{-} &
  Transportation \\ \cline{1-1} \cline{7-14} 
\multicolumn{1}{|c|}{{\cite{38}}} &
  \multicolumn{1}{c|}{} &
  \multicolumn{3}{c|}{} &
  \multicolumn{1}{c|}{\multirow{-3}{*}{Quadcopter}} &
  \multicolumn{1}{c|}{} &
  \multicolumn{1}{c|}{Not reported} &
  \multicolumn{1}{c|}{1kg} &
  \multicolumn{1}{c|}{Box} &
  \multicolumn{1}{c|}{\begin{tabular}[c]{@{}c@{}}Fixed-time ESO based \\ output feedback control\end{tabular}} &
  \multicolumn{1}{c|}{} &
  \multicolumn{1}{c|}{IMU,   GPS} &
   \\ \cline{1-1} \cline{6-6} \cline{8-11} \cline{13-13}
\multicolumn{1}{|c|}{{\cite{40}}} &
  \multicolumn{1}{c|}{} &
  \multicolumn{3}{c|}{\multirow{-3}{*}{-}} &
  \multicolumn{1}{c|}{Hexacopter} &
  \multicolumn{1}{c|}{\multirow{-2}{*}{Outdoor experiment}} &
  \multicolumn{1}{c|}{2.3kg} &
  \multicolumn{1}{c|}{2.2kg} &
  \multicolumn{1}{c|}{Cylindrical object} &
  \multicolumn{1}{c|}{\begin{tabular}[c]{@{}c@{}}Passivity-based  \\ decentralized control\end{tabular}} &
  \multicolumn{1}{c|}{} &
  \multicolumn{1}{c|}{IMU,GNSS} &
  \multirow{-2}{*}{Lift/Transportion} \\ \cline{1-1} \cline{3-11} \cline{13-14} 
\multicolumn{1}{|c|}{{\cite{60}}} &
  \multicolumn{1}{c|}{} &
  \multicolumn{2}{c|}{-} &
  \multicolumn{1}{c|}{\begin{tabular}[c]{@{}c@{}}Wrapping the cable \\ around the object\end{tabular}} &
  \multicolumn{1}{c|}{\begin{tabular}[c]{@{}c@{}}Catenary   \\ quadcopter\end{tabular}} &
  \multicolumn{1}{c|}{Simulation} &
  \multicolumn{1}{c|}{Not reported} &
  \multicolumn{1}{c|}{0.1kg} &
  \multicolumn{1}{c|}{Box} &
  \multicolumn{1}{c|}{Adaptive control} &
  \multicolumn{1}{c|}{\multirow{-3}{*}{-}} &
  \multicolumn{1}{c|}{-} &
  \begin{tabular}[c]{@{}c@{}}Wrap and Pull/Lift/\\ Manipulation/\\ Transportation\end{tabular} \\ \cline{1-1} \cline{3-14} 
\multicolumn{1}{|c|}{{\cite{62}}} &
  \multicolumn{1}{c|}{} &
  \multicolumn{2}{c|}{-} &
  \multicolumn{1}{c|}{Attached grippers} &
  \multicolumn{1}{c|}{Hexacopter} &
  \multicolumn{1}{c|}{Outdoor experiment} &
  \multicolumn{1}{c|}{70gr} &
  \multicolumn{1}{c|}{Not reported} &
  \multicolumn{1}{c|}{Beam} &
  \multicolumn{1}{c|}{Model predictive control} &
  \multicolumn{1}{c|}{RRTc} &
  \multicolumn{1}{c|}{\begin{tabular}[c]{@{}c@{}}Camera, Rangefinder, \\ GPS, Compass\end{tabular}} &
  Transportation \\ \cline{1-1} \cline{3-14} 
\multicolumn{1}{|c|}{{\cite{58}}} &
  \multicolumn{1}{c|}{} &
  \multicolumn{3}{c|}{} &
  \multicolumn{1}{c|}{\begin{tabular}[c]{@{}c@{}}Reconfigurable   \\ quadcopter\end{tabular}} &
  \multicolumn{1}{c|}{\begin{tabular}[c]{@{}c@{}}Physical Simulation \\ with human in the loop\end{tabular}} &
  \multicolumn{2}{c|}{Not reported} &
  \multicolumn{1}{c|}{Rectangular object} &
  \multicolumn{1}{c|}{\begin{tabular}[c]{@{}c@{}}Dual space control approach \\ with tension distribution\end{tabular}} &
  \multicolumn{2}{c|}{-} &
  \begin{tabular}[c]{@{}c@{}}Manipulation/\\ Transportation\end{tabular} \\ \cline{1-1} \cline{6-14} 
\multicolumn{1}{|c|}{{\cite{57}}} &
  \multicolumn{1}{c|}{} &
  \multicolumn{3}{c|}{} &
  \multicolumn{1}{c|}{} &
  \multicolumn{1}{c|}{Outdoor experiment} &
  \multicolumn{1}{c|}{1.282kg} &
  \multicolumn{1}{c|}{0.815kg} &
  \multicolumn{1}{c|}{\begin{tabular}[c]{@{}c@{}}Custom built \\ with hardware setup\end{tabular}} &
  \multicolumn{1}{c|}{Load-leading control} &
  \multicolumn{1}{c|}{} &
  \multicolumn{1}{c|}{GPS,   Compass} &
  \begin{tabular}[c]{@{}c@{}}Lift/Manipulation/\\ Transportation\end{tabular} \\ \cline{1-1} \cline{7-11} \cline{13-14} 
\multicolumn{1}{|c|}{{\cite{59}}} &
  \multicolumn{1}{c|}{} &
  \multicolumn{3}{c|}{} &
  \multicolumn{1}{c|}{} &
  \multicolumn{1}{c|}{Lab experiment} &
  \multicolumn{1}{c|}{Not reported} &
  \multicolumn{1}{c|}{250gr} &
  \multicolumn{1}{c|}{Triangular object} &
  \multicolumn{1}{c|}{\begin{tabular}[c]{@{}c@{}}Distributed vision-based \\ control\end{tabular}} &
  \multicolumn{1}{c|}{} &
  \multicolumn{1}{c|}{\begin{tabular}[c]{@{}c@{}}Monocular camera, \\ IMU\end{tabular}} &
  Manipulation \\ \cline{1-1} \cline{7-11} \cline{13-14} 
\multicolumn{1}{|c|}{{\cite{113}}} &
  \multicolumn{1}{c|}{} &
  \multicolumn{3}{c|}{} &
  \multicolumn{1}{c|}{} &
  \multicolumn{1}{c|}{Gazebo simulation} &
  \multicolumn{1}{c|}{2.02kg} &
  \multicolumn{1}{c|}{1.0kg} &
  \multicolumn{1}{c|}{Cuboid object} &
  \multicolumn{1}{c|}{} &
  \multicolumn{1}{c|}{} &
  \multicolumn{1}{c|}{} &
   \\ \cline{1-1} \cline{7-10}
\multicolumn{1}{|c|}{{\cite{41}}} &
  \multicolumn{1}{c|}{} &
  \multicolumn{3}{c|}{} &
  \multicolumn{1}{c|}{} &
  \multicolumn{1}{c|}{} &
  \multicolumn{1}{c|}{} &
  \multicolumn{1}{c|}{0.4kg} &
  \multicolumn{1}{c|}{Spherical pendulum} &
  \multicolumn{1}{c|}{} &
  \multicolumn{1}{c|}{} &
  \multicolumn{1}{c|}{} &
   \\ \cline{1-1} \cline{9-10}
\multicolumn{1}{|c|}{{\cite{42}}} &
  \multicolumn{1}{c|}{} &
  \multicolumn{3}{c|}{} &
  \multicolumn{1}{c|}{} &
  \multicolumn{1}{c|}{} &
  \multicolumn{1}{c|}{} &
  \multicolumn{1}{c|}{1.5kg} &
  \multicolumn{1}{c|}{} &
  \multicolumn{1}{c|}{} &
  \multicolumn{1}{c|}{} &
  \multicolumn{1}{c|}{} &
  \multirow{-3}{*}{Transportation} \\ \cline{1-1} \cline{9-9} \cline{14-14} 
\multicolumn{1}{|c|}{{\cite{43}}} &
  \multicolumn{1}{c|}{} &
  \multicolumn{3}{c|}{} &
  \multicolumn{1}{c|}{} &
  \multicolumn{1}{c|}{\multirow{-3}{*}{Simulation}} &
  \multicolumn{1}{c|}{} &
  \multicolumn{1}{c|}{0.5kg} &
  \multicolumn{1}{c|}{\multirow{-2}{*}{Rectangular box}} &
  \multicolumn{1}{c|}{} &
  \multicolumn{1}{c|}{} &
  \multicolumn{1}{c|}{} &
  Lift/Transportation \\ \cline{1-1} \cline{7-7} \cline{9-10} \cline{14-14} 
\multicolumn{1}{|c|}{{\cite{44}}} &
  \multicolumn{1}{c|}{} &
  \multicolumn{3}{c|}{} &
  \multicolumn{1}{c|}{} &
  \multicolumn{1}{c|}{} &
  \multicolumn{1}{c|}{\multirow{-4}{*}{0.755kg}} &
  \multicolumn{1}{c|}{0.52kg} &
  \multicolumn{1}{c|}{Rod} &
  \multicolumn{1}{c|}{\multirow{-5}{*}{Geometric control}} &
  \multicolumn{1}{c|}{} &
  \multicolumn{1}{c|}{} &
  Stabilization \\ \cline{1-1} \cline{8-11} \cline{14-14} 
\multicolumn{1}{|c|}{{\cite{68}}} &
  \multicolumn{1}{c|}{} &
  \multicolumn{3}{c|}{} &
  \multicolumn{1}{c|}{} &
  \multicolumn{1}{c|}{} &
  \multicolumn{1}{c|}{400gr} &
  \multicolumn{1}{c|}{400gr} &
  \multicolumn{1}{c|}{Cubic object} &
  \multicolumn{1}{c|}{\begin{tabular}[c]{@{}c@{}}Incremental Nonlinear Dynamic \\ Inversion controller \\ and robust formation\end{tabular}} &
  \multicolumn{1}{c|}{} &
  \multicolumn{1}{c|}{} &
  \begin{tabular}[c]{@{}c@{}}Lift/Transportation/\\ Manipulation\end{tabular} \\ \cline{1-1} \cline{8-11} \cline{14-14} 
\multicolumn{1}{|c|}{{\cite{69}}} &
  \multicolumn{1}{c|}{} &
  \multicolumn{3}{c|}{} &
  \multicolumn{1}{c|}{} &
  \multicolumn{1}{c|}{} &
  \multicolumn{1}{c|}{1kg} &
  \multicolumn{1}{c|}{0.9kg} &
  \multicolumn{1}{c|}{Custom} &
  \multicolumn{1}{c|}{\begin{tabular}[c]{@{}c@{}}Distributed trajectory \\ optimization control\end{tabular}} &
  \multicolumn{1}{c|}{} &
  \multicolumn{1}{c|}{} &
  Transportation \\ \cline{1-1} \cline{8-11} \cline{14-14} 
\multicolumn{1}{|c|}{{\cite{67}}} &
  \multicolumn{1}{c|}{} &
  \multicolumn{3}{c|}{} &
  \multicolumn{1}{c|}{} &
  \multicolumn{1}{c|}{\multirow{-4}{*}{Lab experiment}} &
  \multicolumn{1}{c|}{500gr} &
  \multicolumn{1}{c|}{575gr} &
  \multicolumn{1}{c|}{Aluminum bar} &
  \multicolumn{1}{c|}{Adaptive control} &
  \multicolumn{1}{c|}{} &
  \multicolumn{1}{c|}{} &
   \\ \cline{1-1} \cline{7-11}
\multicolumn{1}{|c|}{{\cite{32}}} &
  \multicolumn{1}{c|}{} &
  \multicolumn{3}{c|}{\multirow{-12}{*}{-}} &
  \multicolumn{1}{c|}{} &
  \multicolumn{1}{c|}{Simulation} &
  \multicolumn{1}{c|}{1kg} &
  \multicolumn{1}{c|}{0.9kg} &
  \multicolumn{1}{c|}{Beam} &
  \multicolumn{1}{c|}{Admittance controller} &
  \multicolumn{1}{c|}{} &
  \multicolumn{1}{c|}{\multirow{-9}{*}{-}} &
  \multirow{-2}{*}{\begin{tabular}[c]{@{}c@{}}Manipulation/\\ Transportation\end{tabular}} \\ \cline{1-1} \cline{3-5} \cline{7-11} \cline{13-14} 
\multicolumn{1}{|c|}{{\cite{61}}} &
  \multicolumn{1}{c|}{} &
  \multicolumn{2}{c|}{-} &
  \multicolumn{1}{c|}{Carabiner-harness} &
  \multicolumn{1}{c|}{} &
  \multicolumn{1}{c|}{Outdoor experiment} &
  \multicolumn{1}{c|}{1kg} &
  \multicolumn{1}{c|}{3kg} &
  \multicolumn{1}{c|}{\begin{tabular}[c]{@{}c@{}}Custom built \\ with hardware setup\end{tabular}} &
  \multicolumn{1}{c|}{Cascaded PID} &
  \multicolumn{1}{c|}{} &
  \multicolumn{1}{c|}{GPS} &
  \begin{tabular}[c]{@{}c@{}}Lift/Manipulation/\\ Transportation\end{tabular} \\ \cline{1-1} \cline{3-5} \cline{7-11} \cline{13-14} 
\multicolumn{1}{|c|}{{\cite{106}}} &
  \multicolumn{1}{c|}{} &
  \multicolumn{3}{c|}{-} &
  \multicolumn{1}{c|}{} &
  \multicolumn{1}{c|}{Lab experiment} &
  \multicolumn{1}{c|}{1.43kg} &
  \multicolumn{1}{c|}{0.355kg} &
  \multicolumn{1}{c|}{Cargo} &
  \multicolumn{1}{c|}{\begin{tabular}[c]{@{}c@{}}Nonlinear hierarchical \\ controller\end{tabular}} &
  \multicolumn{1}{c|}{\multirow{-13}{*}{-}} &
  \multicolumn{1}{c|}{-} &
  Cargo delivery \\ \cline{1-1} \cline{3-5} \cline{7-14} 
\multicolumn{1}{|c|}{{\cite{108}}} &
  \multicolumn{1}{c|}{} &
  \multicolumn{2}{c|}{3-DOF arm (x1)} &
  \multicolumn{1}{c|}{Gripper} &
  \multicolumn{1}{c|}{} &
  \multicolumn{1}{c|}{Simulation} &
  \multicolumn{1}{c|}{\begin{tabular}[c]{@{}c@{}}1.2kg   \\ (with manipulator)\end{tabular}} &
  \multicolumn{1}{c|}{0.1kg} &
  \multicolumn{1}{c|}{Slung load} &
  \multicolumn{1}{c|}{\begin{tabular}[c]{@{}c@{}}Null-Space-Based   \\ Adaptive Control\end{tabular}} &
  \multicolumn{2}{c|}{-} &
  Transportation \\ \cline{1-1} \cline{3-5} \cline{7-14} 
\multicolumn{1}{|c|}{{\cite{109}}} &
  \multicolumn{1}{c|}{} &
  \multicolumn{3}{c|}{} &
  \multicolumn{1}{c|}{} &
  \multicolumn{1}{c|}{Outdoor experiment} &
  \multicolumn{1}{c|}{1.46kg} &
  \multicolumn{1}{c|}{1.5kg} &
  \multicolumn{1}{c|}{\begin{tabular}[c]{@{}c@{}}Rectangular   \\ aluminum frame\end{tabular}} &
  \multicolumn{1}{c|}{Collaborative Control} &
  \multicolumn{1}{c|}{-} &
  \multicolumn{1}{c|}{GPS,   IMU} &
  \begin{tabular}[c]{@{}c@{}}Lift/Transportation/\\ Land\end{tabular} \\ \cline{1-1} \cline{7-14} 
\multicolumn{1}{|c|}{{\cite{114}}} &
  \multicolumn{1}{c|}{} &
  \multicolumn{3}{c|}{} &
  \multicolumn{1}{c|}{} &
  \multicolumn{1}{c|}{Lab experiment} &
  \multicolumn{1}{c|}{0.67kg} &
  \multicolumn{1}{c|}{0.4kg} &
  \multicolumn{1}{c|}{Spherical object} &
  \multicolumn{1}{c|}{Passivity-based control} &
  \multicolumn{2}{c|}{} &
   \\ \cline{1-1} \cline{7-11}
\multicolumn{1}{|c|}{{\cite{115}}} &
  \multicolumn{1}{c|}{} &
  \multicolumn{3}{c|}{\multirow{-3}{*}{-}} &
  \multicolumn{1}{c|}{} &
  \multicolumn{1}{c|}{Simulation} &
  \multicolumn{1}{c|}{1.15kg} &
  \multicolumn{1}{c|}{0.5kg} &
  \multicolumn{1}{c|}{Spherical object} &
  \multicolumn{1}{c|}{\begin{tabular}[c]{@{}c@{}}Sliding Mode-Adaptive \\ PID control\end{tabular}} &
  \multicolumn{2}{c|}{\multirow{-2}{*}{-}} &
  \multirow{-2}{*}{Transportation} \\ \cline{1-1} \cline{3-5} \cline{7-14} 
\multicolumn{1}{|c|}{{\cite{116}}} &
  \multicolumn{1}{c|}{} &
  \multicolumn{1}{c|}{-} &
  \multicolumn{2}{c|}{Spherical Joint} &
  \multicolumn{1}{c|}{} &
  \multicolumn{1}{c|}{} &
  \multicolumn{1}{c|}{\begin{tabular}[c]{@{}c@{}}1.18kg and \\ 80.4kg structure\end{tabular}} &
  \multicolumn{1}{c|}{2.5kg} &
  \multicolumn{1}{c|}{Box} &
  \multicolumn{1}{c|}{-} &
  \multicolumn{2}{c|}{} &
  Lift/Transportation \\ \cline{1-1} \cline{3-5} \cline{8-14} 
\multicolumn{1}{|c|}{{\cite{117}}} &
  \multicolumn{1}{c|}{} &
  \multicolumn{3}{c|}{} &
  \multicolumn{1}{c|}{} &
  \multicolumn{1}{c|}{\multirow{-2}{*}{Lab experiment}} &
  \multicolumn{2}{c|}{Not reported} &
  \multicolumn{1}{c|}{Rectangular bar} &
  \multicolumn{1}{c|}{LQR} &
  \multicolumn{1}{c|}{} &
  \multicolumn{1}{c|}{\begin{tabular}[c]{@{}c@{}}IMU, Camera, \\ Ultrasound sensor\end{tabular}} &
   \\ \cline{1-1} \cline{7-11} \cline{13-13}
\multicolumn{1}{|c|}{{\cite{119}}} &
  \multicolumn{1}{c|}{} &
  \multicolumn{3}{c|}{} &
  \multicolumn{1}{c|}{} &
  \multicolumn{1}{c|}{Outdoor experiment} &
  \multicolumn{1}{c|}{73gr} &
  \multicolumn{1}{c|}{30gr} &
  \multicolumn{1}{c|}{Circular object} &
  \multicolumn{1}{c|}{\begin{tabular}[c]{@{}c@{}}Linear controller \\ with EKF-based estimator\end{tabular}} &
  \multicolumn{1}{c|}{\multirow{-2}{*}{-}} &
  \multicolumn{1}{c|}{IMU} &
   \\ \cline{1-1} \cline{7-13}
\multicolumn{1}{|c|}{{\cite{120}}} &
  \multicolumn{1}{c|}{} &
  \multicolumn{3}{c|}{} &
  \multicolumn{1}{c|}{} &
  \multicolumn{1}{c|}{} &
  \multicolumn{1}{c|}{Not reported} &
  \multicolumn{1}{c|}{0.39kg} &
  \multicolumn{1}{c|}{Circular object} &
  \multicolumn{1}{c|}{\begin{tabular}[c]{@{}c@{}}Estimation-based formation \\ tracking control\end{tabular}} &
  \multicolumn{1}{c|}{Formation Planning} &
  \multicolumn{1}{c|}{-} &
   \\ \cline{1-1} \cline{8-13}
\multicolumn{1}{|c|}{{\cite{121}}} &
  \multicolumn{1}{c|}{} &
  \multicolumn{3}{c|}{} &
  \multicolumn{1}{c|}{} &
  \multicolumn{1}{c|}{} &
  \multicolumn{1}{c|}{800gr} &
  \multicolumn{1}{c|}{300gr} &
  \multicolumn{1}{c|}{Circular object} &
  \multicolumn{1}{c|}{\begin{tabular}[c]{@{}c@{}}Backstepping force controller \\ and force disturbance observer\end{tabular}} &
  \multicolumn{1}{c|}{-} &
  \multicolumn{1}{c|}{IMU} &
  \multirow{-4}{*}{Transportation} \\ \cline{1-1} \cline{8-14} 
\multicolumn{1}{|c|}{{\cite{123}}} &
  \multicolumn{1}{c|}{} &
  \multicolumn{3}{c|}{} &
  \multicolumn{1}{c|}{} &
  \multicolumn{1}{c|}{} &
  \multicolumn{1}{c|}{Not reported} &
  \multicolumn{1}{c|}{232gr} &
  \multicolumn{1}{c|}{Triangular object} &
  \multicolumn{1}{c|}{\begin{tabular}[c]{@{}c@{}}Nonlinear Model Predictive \\ Control\end{tabular}} &
  \multicolumn{2}{c|}{} &
  \begin{tabular}[c]{@{}c@{}}Manipulation/\\ Transportation\end{tabular} \\ \cline{1-1} \cline{8-11} \cline{14-14} 
\multicolumn{1}{|c|}{{\cite{118}}} &
  \multicolumn{1}{c|}{} &
  \multicolumn{3}{c|}{} &
  \multicolumn{1}{c|}{} &
  \multicolumn{1}{c|}{} &
  \multicolumn{2}{c|}{Not reported} &
  \multicolumn{1}{c|}{Bar} &
  \multicolumn{1}{c|}{\begin{tabular}[c]{@{}c@{}}Backstepping (leader), \\ Switching Controller (follower)\end{tabular}} &
  \multicolumn{2}{c|}{} &
  Transportation \\ \cline{1-1} \cline{8-11} \cline{14-14} 
\multicolumn{1}{|c|}{{\cite{124}}} &
  \multicolumn{1}{c|}{\multirow{-35}{*}{Fixed}} &
  \multicolumn{3}{c|}{\multirow{-7}{*}{-}} &
  \multicolumn{1}{c|}{\multirow{-25}{*}{Quadcopter}} &
  \multicolumn{1}{c|}{\multirow{-5}{*}{Lab experiment}} &
  \multicolumn{1}{c|}{\begin{tabular}[c]{@{}c@{}}0.831kg   \\ and 0.832kg\end{tabular}} &
  \multicolumn{1}{c|}{\begin{tabular}[c]{@{}c@{}}0.248kg   \\ and  0.252kg\end{tabular}} &
  \multicolumn{1}{c|}{Steel pipe} &
  \multicolumn{1}{c|}{\begin{tabular}[c]{@{}c@{}}Force Coordination \\ Control\end{tabular}} &
  \multicolumn{2}{c|}{\multirow{-3}{*}{-}} &
  \begin{tabular}[c]{@{}c@{}}Hovering/   \\ Transportation\end{tabular} \\ \hline
\end{tabular}%
}
\end{table}

\bibliographystyle{unsrt}  
\bibliography{coop}

\end{document}